\documentclass[10pt,a4paper,oneside,notitlepage,fleqn,reqno,english,table]{article}

\usepackage[hidelinks]{hyperref} 

\usepackage{multicol} 

\usepackage{longtable} 
\usepackage{multirow} 
\usepackage{rotating} 
\usepackage[cp1252]{inputenc} 
\usepackage{lscape} 
\usepackage{url}    
\usepackage[section]{placeins}  
\usepackage{afterpage} 
\usepackage{capt-of}        
\usepackage{dirtytalk} 

\usepackage[T1]{fontenc}
\usepackage{tabularx, booktabs}   
\usepackage{makecell} 
\usepackage{array} 
\usepackage{xcolor} 
\usepackage{xfrac} 

\usepackage{tikz}  
\usetikzlibrary{shapes,arrows,arrows.meta,calc,shadows,fit,intersections,datavisualization.formats.functions}
\tikzstyle{block}        = [draw, fill=blue!30, rectangle, text centered, minimum height=3em, minimum width=6em] 
\tikzstyle{blockred}     = [draw, fill=red!30, rectangle, text centered, minimum height=3em, minimum width=6em] 
\tikzstyle{blockyellow}  = [draw, fill=yellow!30, rectangle, text centered, minimum height=3em, minimum width=6em] 
\tikzstyle{blockyellow1} = [draw, fill=yellow!15, rectangle, text centered, minimum height=3em, minimum width=6em] 
\tikzstyle{blockyellow2} = [draw, fill=yellow!30, rectangle, text centered, minimum height=3em, minimum width=6em] 
\tikzstyle{blockyellow3} = [draw, fill=yellow!45, rectangle, text centered, minimum height=3em, minimum width=6em] 
\tikzstyle{blockyellow4} = [draw, fill=yellow!60, rectangle, text centered, minimum height=3em, minimum width=6em] 
\tikzstyle{blockgreen}   = [draw, fill=green!30, rectangle, text centered, minimum height=3em, minimum width=6em] 
\tikzstyle{blockgreen1}  = [draw, fill=green!15, rectangle, text centered, minimum height=3em, minimum width=6em] 
\tikzstyle{blockgreen2}  = [draw, fill=green!30, rectangle, text centered, minimum height=3em, minimum width=6em] 
\tikzstyle{blockgreen3}  = [draw, fill=green!45, rectangle, text centered, minimum height=3em, minimum width=6em] 
\tikzstyle{blockgreen4}  = [draw, fill=green!60, rectangle, text centered, minimum height=3em, minimum width=6em] 
\tikzstyle{blockgrey1}   = [draw, fill=black!8,  rectangle, text centered, minimum height=3em, minimum width=6em] 
\tikzstyle{blockgrey2}   = [draw, fill=black!16, rectangle, text centered, minimum height=3em, minimum width=6em] 
\tikzstyle{blockgrey3}   = [draw, fill=black!24, rectangle, text centered, minimum height=3em, minimum width=6em] 
\tikzstyle{blockgrey4}   = [draw, fill=black!32, rectangle, text centered, minimum height=3em, minimum width=6em] 
\tikzstyle{blockgrey5}   = [draw, fill=black!40, rectangle, text centered, minimum height=3em, minimum width=6em] 
\tikzstyle{blocknofill}  = [draw=black!50, line width=1.5pt, rectangle, rounded corners, text centered, minimum height=3em, minimum width=6em]
\tikzstyle{blockhigh}    = [draw, fill=blue!20, rectangle, text centered, minimum height=6em, minimum width=6em]
\tikzstyle{noblock}      = [draw=black!50, line width=1.5pt, rectangle, rounded corners, text centered, minimum height=3em, minimum width=6em]
\tikzstyle{sum}          = [draw, fill=blue!20, circle, node distance=1cm]
\tikzstyle{sumrest}      = [draw, fill=black, circle, radius=0.5cm]
\tikzstyle{pinstyle}     = [pin edge={to-,thin,black}]
\tikzstyle{title}        = [text centered]
\pgfdeclarelayer{background}
\pgfdeclarelayer{foreground}
\pgfsetlayers{background,main,foreground}

\usepackage{pgfplots}
\pgfplotsset{compat=1.13}
\pgfplotsset{colormap/bluered}
\usepackage{pgfplotstable}
\usepgfplotslibrary{polar}
\usepgfplotslibrary{colormaps}
\usepgfplotslibrary{external} 

\graphicspath{{figs/}} 					
\usepackage[parfill]{parskip} 					

\usepackage{amsmath} 			
\usepackage{amsfonts}			
\usepackage{mathtools}			
\everymath{\displaystyle}		
\usepackage{amssymb}  			
\usepackage{nicefrac}  			

\usepackage{tikz}  
\usetikzlibrary{shapes,arrows,calc,shadows,fit,intersections,datavisualization.formats.functions}
\tikzstyle{block}        = [draw, fill=blue!30, rectangle, text centered, minimum height=3em, minimum width=6em] 
\tikzstyle{title}        = [text centered]
\pgfdeclarelayer{background}
\pgfdeclarelayer{foreground}
\pgfsetlayers{background,main,foreground}

\usepackage{pgfplots}
\pgfplotsset{compat=1.13}
\pgfplotsset{colormap/bluered}
\usepackage{pgfplotstable}
\usepgfplotslibrary{polar}
\usepgfplotslibrary{colormaps}
\usepgfplotslibrary{external}

\usepackage{graphicx, color} 	

\allowdisplaybreaks 

\newcommand{\sss}   {\scriptscriptstyle}    

\newcommand{\nm}   		[1] {\ensuremath{\mathrm{#1}}} 									
\newcommand{\neweq}     [2] {\begin{equation} \mathrm{#1}\label{#2} \end{equation}} 	

\newcommand{\CC}		{{C\nolinebreak[4]\hspace{-.05em}\raisebox{.4ex}{\tiny\bf ++}}}

\renewcommand{\vec}		[1] {\mbox{\boldmath{\ensuremath{\mathrm{#1}}}}}	
\newcommand{\lrp}       [1] {\left(#1\right)}								
\newcommand{\lrsb}      [1] {\left[#1\right]}								
\newcommand{\lrb}       [1] {\left\{#1\right\}}								

\newcommand{\muj}        [1] {\mu_{{#1}j}}
\newcommand{\sigmaj}     [1] {\sigma_{{#1}j}}
\newcommand{\maxj}       [1] {{\zeta}_{\lvert {#1} \rvert j}}
\newcommand{\XENDj}      [1] {{#1}_{{\sss{END}}j}}

\newcommand{\mumu}       [1] {\mu_{\mu {#1}}}
\newcommand{\musigma}    [1] {\mu_{\sigma {#1}}}
\newcommand{\mumax}      [1] {\mu_{\lvert{{\zeta} \lvert {#1} \rvert}\rvert}}

\newcommand{\sigmamu}    [1] {\sigma_{\mu {#1}}}
\newcommand{\sigmasigma} [1] {\sigma_{\sigma {#1}}}
\newcommand{\sigmamax}   [1] {\sigma_{\lvert{{\zeta} \lvert {#1} \rvert}\rvert}}

\newcommand{\maxmu}      [1] {{\zeta}_{\lvert \mu {#1} \rvert}}
\newcommand{\maxsigma}   [1] {{\zeta}_{\lvert \sigma {#1} \rvert}}
\newcommand{\maxmax}     [1] {{\zeta}_{\lvert {\zeta} \lvert {#1} \rvert \rvert}}

\newcommand{\muEND}      [1] {\mu_{{\sss{END}} {#1}}}
\newcommand{\sigmaEND}   [1] {\sigma_{{\sss{END}} {#1}}}
\newcommand{\maxEND}     [1] {{\zeta}_{{\sss{END}}\lvert {#1} \rvert}}

\newcommand{\Deltat}     	{\Delta t}
\newcommand{\DeltatTRUTH}	{\Delta t_{\sss TRUTH}}
\newcommand{\DeltatSENSED}  {\Delta t_{\sss SENSED}}
\newcommand{\DeltatEST}  	{\Delta t_{\sss EST}}
\newcommand{\DeltatCNTR}  	{\Delta t_{\sss CNTR}}

\newcommand{\deltaCNTR}  	{\vec{\delta}_{\sss CNTR}}		
\newcommand{\deltaTARGET}  	{\vec{\delta}_{\sss TARGET}}	    
\newcommand{\deltaTARGETone}  	{\vec{\delta}_{\sss TARGET,1}}
\newcommand{\deltaTARGETtwo}  	{\vec{\delta}_{\sss TARGET,2}}
\newcommand{\deltaTARGETthr}  	{\vec{\delta}_{\sss TARGET,3}}
\newcommand{\deltaTARGETfou}  	{\vec{\delta}_{\sss TARGET,4}}
\newcommand{\deltaTARGETfiv}  	{\vec{\delta}_{\sss TARGET,5}}
\newcommand{\deltaTARGETsix}  	{\vec{\delta}_{\sss TARGET,6}}

\newcommand{\deltaTARGETfourteen}  	{\vec{\delta}_{\sss TARGET,14}}
\newcommand{\deltaTARGETfifteen}  	{\vec{\delta}_{\sss TARGET,15}}
\newcommand{\deltaTARGETsixteen}  	{\vec{\delta}_{\sss TARGET,16}}
\newcommand{\deltaTARGETseventeen} 	{\vec{\delta}_{\sss TARGET,17}}

\newcommand{\deltaT}  		{\delta_{\sss T}}            	
\newcommand{\deltaE}		{\delta_{\sss E}}            	
\newcommand{\deltaA}		{\delta_{\sss A}}				
\newcommand{\deltaR}		{\delta_{\sss R}}				
\newcommand{\deltaTARGETT}	{\delta_{\sss{TARGET,T}}}		    
\newcommand{\deltaTARGETE}	{\delta_{\sss{TARGET,E}}}		    
\newcommand{\deltaTARGETA}	{\delta_{\sss{TARGET,A}}}		    
\newcommand{\deltaTARGETR}	{\delta_{\sss{TARGET,R}}}		    
\newcommand{\deltaTRG}  	{\delta_{\sss{TRG}}}			

\newcommand{\xvec}   			{\vec x}
\newcommand{\xveczero}	 		{\vec x_0}
\newcommand{\xvecestzero}		{\hat{\vec x}_0}
\newcommand{\xvecdot}			{\dot {\vec x}}
\newcommand{\xvecest}	  		{\hat{\vec x}}
\newcommand{\xvectilde} 		{\widetilde{\vec x}}

\newcommand{\xTRUTH}			{\vec x_{\sss TRUTH}}
\newcommand{\xSENSED}  			{\vec x_{\sss SENSED}}
\newcommand{\xEST}  			{\vec x_{\sss EST}}
\newcommand{\xREF}	  			{\vec x_{\sss REF}}

\newcommand{\xEgdt}      {\vec x_{\sss GDT}}

\newcommand{\gc}     	  {\vec g_c}

\newcommand{\vB} 	   			{\vec v^{\sss B}}

\newcommand{\vNtilde}			{\widetilde{\vec v}^{\sss N}}

\newcommand{\vtas}          {v_{\sss TAS}}
\newcommand{\vtasINI}       {v_{\sss TAS,INI}}
\newcommand{\vtasEND}       {v_{\sss TAS,END}}
\newcommand{\vtastilde}     {\widetilde{v}_{\sss TAS}}

\newcommand{\vwindINI}     	{v_{\sss WIND,INI}}
\newcommand{\vwindEND}     	{v_{\sss WIND,END}}
\newcommand{\vWINDINI}     	{\vec v_{\sss WIND,INI}}
\newcommand{\vWINDEND}     	{\vec v_{\sss WIND,END}}

\newcommand{\wNBB}			{\vec \omega_{\sss NB}^{\sss B}}

\newcommand{\wIBB}				{\vec \omega_{\sss IB}^{\sss B}}

\newcommand{\wIBBtilde}			{\widetilde{\vec \omega}_{\sss IB}^{\sss B}}

\newcommand{\qNB}			{\vec q_{\sss NB}}

\newcommand{\gammaTAS}  	  	{\gamma_{\sss TAS}}

\newcommand{\chiWINDINI}       	{\chi_{\sss WIND,INI}}
\newcommand{\chiWINDEND}       	{\chi_{\sss WIND,END}}

\newcommand{\chiINI}            {\chi_{\sss INI}}
\newcommand{\chiEND}            {\chi_{\sss END}}

\newcommand{\xiTURN}            {\xi_{\sss TURN}}
\newcommand{\xiTURNi}           {\xi_{{\sss TURN},i}}
\newcommand{\gammaTASCLIMB}     {\gamma_{\sss TAS,CLIMB}}

\newcommand{\DeltatOPTAS}       {\Deltat_{\sss OP,TAS}}
\newcommand{\DeltatOPHp}        {\Deltat_{\sss OP,Hp}}

\newcommand{\tTURN}				{t_{\sss TURN}}
\newcommand{\DeltatTURNi}		{\Delta t_{{\sss TURN},i}}
\newcommand{\tGNSS}				{t_{\sss GNSS}}
\newcommand{\tEND}				{t_{\sss END}}
\newcommand{\tINIDeltaT}		{t_{\sss INI,\Delta T}}
\newcommand{\tENDDeltaT}		{t_{\sss END,\Delta T}}
\newcommand{\tINIDeltap}		{t_{\sss INI,\Delta p}}
\newcommand{\tENDDeltap}		{t_{\sss END,\Delta p}}
\newcommand{\tINIWIND}		    {t_{\sss INI,WIND}}
\newcommand{\tENDWIND}	   	    {t_{\sss END,WIND}}

\newcommand{\Hp}            {H_{\sss P}}
\newcommand{\HpINI}         {H_{\sss P,INI}}
\newcommand{\HpEND}         {H_{\sss P,END}}

\newcommand{\DeltaT}        {\Delta T}
\newcommand{\Deltap}        {\Delta p}

\newcommand{\DeltaTINI}     {\Delta T_{\sss INI}}
\newcommand{\DeltaTEND}     {\Delta T_{\sss END}}

\newcommand{\DeltapINI}     {\Delta p_{\sss INI}}
\newcommand{\DeltapEND}     {\Delta p_{\sss END}}

\newcommand{\Tzero}         {T_{\sss 0}}                 
\newcommand{\pzero}         {p_{\sss 0}}                 

\newcommand{\RE}     	    {R_{\sss E}}   				

\newcommand{\fIBBtilde}			{\widetilde{\vec f}_{\sss IB}^{\sss B}}

\newcommand{\nEND}				{n_{\sss END}}
\newcommand{\mun}        [1] {\mu_{{#1}n}}
\newcommand{\sigman}     [1] {\sigma_{{#1}n}}

\newcommand{\BBtilde}		{\widetilde{\vec B}^{\sss B}}

\newcommand{\nEX}           {n_{\sss EX}}
\newcommand{\seed} 			{\Upsilon_{j}}

\newcommand{\seedACC} 		{\upsilon_{j,\sss ACC}}
\newcommand{\seedGYR}		{\upsilon_{j,\sss GYR}}
\newcommand{\seedMAG}		{\upsilon_{j,\sss MAG}}
\newcommand{\seedPLAT}		{\upsilon_{j,\sss PLAT}}
\newcommand{\seedOSP}		{\upsilon_{j,\sss OSP}}
\newcommand{\seedOAT}		{\upsilon_{j,\sss OAT}}
\newcommand{\seedGNSS}		{\upsilon_{j,\sss GNSS}}
\newcommand{\seedTAS}		{\upsilon_{j,\sss TAS}}
\newcommand{\seedAOA}		{\upsilon_{j,\sss AOA}}
\newcommand{\seedAOS}		{\upsilon_{j,\sss AOS}}
\newcommand{\seedCAM}		{\upsilon_{j,\sss CAM}}
\newcommand{\seedWEATHER}	{\upsilon_{j,\sss WEATHER}}
\newcommand{\seedWIND}		{\upsilon_{j,\sss WIND}}
\newcommand{\seedTURB}		{\upsilon_{j,\sss TURB}}
\newcommand{\seedMISSION}	{\upsilon_{j,\sss MISSION}}
\newcommand{\seedGEO}		{\upsilon_{j,\sss GEO}}
\newcommand{\seedALIGN}		{\upsilon_{j,\sss ALIGN}}

\begin{document}

\title{Stochastic High Fidelity Simulation and Scenarios for Testing of Fixed Wing Autonomous GNSS-Denied Navigation Algorithms}
\author{Eduardo Gallo\footnote{The author holds a MSc in Aerospace Engineering by the Polytechnic University of Madrid and has twenty-two years of experience working in aircraft trajectory prediction, modeling, and flight simulation. He is currently a Senior Trajectory Prediction and Aircraft Performance Engineer at Boeing Research \& Technology Europe (BR\&TE), although he is publishing this article in his individual capacity and time as part of his PhD thesis titled ``Autonomous Unmanned Air Vehicle GNSS-Denied Navigation'', advised by Dr. Antonio Barrientos within the Centre for Automation and Robotics of the Polytechnic University of Madrid.} \footnote{Contact: edugallo@yahoo.com, \url{https://orcid.org/0000-0002-7397-0425}}}
\date{February 2021}
\maketitle


\section*{Abstract}

Autonomous unmanned aerial vehicle (UAV) inertial navigation exhibits an extreme dependency on the availability of global navigation satellite systems (GNSS) signals, without which it incurs in a slow but unavoidable position drift that may ultimately lead to the loss of the platform if the GNSS signals are not restored or the aircraft does not reach a location from which it can be recovered by remote control. This article describes an stochastic high fidelity simulation of the flight of a fixed wing low SWaP (size, weight, and power) autonomous UAV in turbulent and varying weather intended to test and validate the GNSS-Denied performance of different navigation algorithms. Its open-source \nm{\CC} implementation is available in \cite{Gallo2020_simulation}. 

Onboard sensors include accelerometers, gyroscopes, magnetometers, a Pitot tube, an air data system, a GNSS receiver, and a digital camera, so the simulation is valid for inertial, visual, and visual inertial navigation systems. Two scenarios involving the loss of GNSS signals are considered: the first represents the challenges involved in aborting the mission and heading towards a remote recovery location while experiencing varying weather, and the second models the continuation of the mission based on a series of closely spaced bearing changes.

All simulation modules have been modeled with as few simplifications as possible to increase the realism of the results. While the implementation of the aircraft performances and its control system is deterministic, that of all other modules, including the mission, sensors, weather, wind, turbulence, and initial estimations, is fully stochastic. This enables a robust evaluation of each proposed navigation system by means of Monte-Carlo simulations that rely on a high number of executions of both scenarios. 


\section{Outline}\label{sec:Outline}

This article begins with a short introduction to GNSS-Denied navigation, followed by an explanation of the approach taken to introduce randomness into the simulation by means of multiple stochastic parameters. The next section focuses on the high-fidelity modeling of the different simulation modules (with the exception of the navigation system), which is followed by a detailed description of both scenarios and the definition of the metrics employed for evaluation. Future articles will propose and evaluate different navigation systems based on the stochastic high fidelity simulation described in this article.


\section{GNSS-Denied Navigation}\label{sec:GNSS-Denied}

UAVs can be classified into remotely controlled aircraft, in which the ground human operator continuously provides control commands to the platform, and autonomous aircraft, which rely on onboard computers to execute the previously uploaded mission objectives. In the case of remotely piloted platforms, the UAV also employs the communications channel to provide the ground with its current status (position, velocity, attitude, etc.) and in some cases photos or video, on which the ground operator relies to continuously generate the control instructions. Autonomous aircraft can also employ the communications channel to provide information to the ground, which may decide to update the mission objectives based on it, but as its own name implies, they usually operate without any kind of communication with its operator. They just continue executing its mission until the flight concludes or communications are restored.

The number, variety, and applications of UAVs have grown exponentially in the last few years, and the trend is expected to continue in the future. This is particularly true in the case of low SWaP vehicles because their reduced cost makes them suitable for a wide range of applications, both civil and military. A significant percentage of these vehicles are capable of operating autonomously. With small variations, these platforms rely on a suite of sensors that continuously provides noisy data about the airframe state, a navigation algorithm to estimate the aircraft pose (position plus attitude), and a control system that, based on the navigation output, adjusts the aircraft control mechanisms to successfully execute the preloaded mission.

This article focuses on fixed wing autonomous platforms, which are generally equipped with a GNSS receiver, accelerometers, gyroscopes, magnetometers, an air data system, a Pitot tube, and one or multiple cameras\footnote{A single camera is employed in the simulation.}. The combination of accelerometers and gyroscopes is known as the Inertial Measurement Unit (IMU). The errors introduced by all sensors grow significantly as their SWaP decreases, in particular in the case of the IMU. The recent introduction of solid state accelerometers and gyroscopes has dramatically improved the performance of low SWaP IMUs, with new models showing significant improvements when compared to those fabricated only a few years ago. The problem of noisy measurement readings is compounded in the case of low SWaP vehicles by the low mass of the platforms, which results in less inertia and hence more high frequency accelerations and rotations caused by the atmospheric turbulence.

Aircraft navigation has traditionally relied on the measurements provided by accelerometers, gyroscopes, and magnetometers, incurring in an slow but unbounded position drift that could only be stopped by triangulation with the use of external navigation (radio) aids. More recently, the introduction of satellite navigation has completely removed the position drift and enabled autonomous \emph{inertial navigation} in low SWaP platforms \cite{Farrell2008, Groves2008, Chatfield1997}. On the negative side, low SWaP inertial navigation exhibits an extreme dependency on the availability of GNSS signals. If not present, inertial systems rely on dead reckoning, which results in velocity and position drift, with the aircraft slowly but steadily deviating from its intended route.

The availability of GNSS signals cannot be guaranteed by any means. In addition to the (unlikely) event of one or various GNSS satellites ceasing to broadcast (voluntarily or not), the GNSS signals can be accidentally blocked by nearby objects (mountains, buildings, trees), corrupted by their own reflections in those same objects, or maliciously interfered with by broadcasting a different signal in the same frequency. Any of the above results in what is known as \emph{GNSS-Denied navigation}. In that event, the vehicle is unable to fly its intended route or even return to a safe recovery location, which leads to the uncontrolled loss of the airframe if the GNSS signal is not recovered before the aircraft runs out of fuel (or battery in case of electric vehicles). \cite{Tippitt2020} provides a summary of the challenges of GNSS-Denied navigation and the research efforts intended to improve its performance.

The extreme dependency on GNSS availability is not only one of the main impediments for the introduction of small UAVs in civil airspace, where it is not acceptable to have uncontrolled vehicles causing personal or material damage, but it also presents a significant drawback for military applications, as a single hull loss may compromise the onboard technology. At this time there are no comprehensive solutions to the operation of low SWaP autonomous UAVs in GNSS-Denied scenarios, although the use of onboard cameras to provide an additional relative pose measurement seems to be one of the most promising routes. Bigger and more expensive UAVs, this is, with less stringent SWaP requirements, can rely to some degree on much more accurate IMUs (at the expense of SWaP), and additional communications equipment to overcome this problem, but for most autonomous UAVs, the permanent loss of the GNSS signal is equivalent to losing the airframe in an uncontrolled way.

\emph{Visual navigation} relies on the images taken by one or more onboard cameras to estimate the variation with time of the aircraft position and attitude \cite{Scaramuzza2011, Fraundorfer2012}. Based  on \emph{visual Odometry} (VO), it also incurs in a slow but unbounded position drift that can only be eliminated if the vehicle revisits a given zone for a second time, a technique known as \emph{simultaneous localization and mapping} (SLAM). If this is not the case, the drift can also be removed through \emph{image registration}, in which the images are compared with those stored in a database. Inertial and visual navigation techniques can be integrated to improve their combined GNSS-Denied navigation capabilities, resulting in a \emph{visual inertial navigation system} or \emph{visual inertial odometry} (VIO). 


\section{Stochastic Simulation}\label{sec:StochasticSimulation}

The trajectory flown by an aircraft is the result of a combination of physical and electronic processes, some deterministic and others stochastic\footnote{The aircraft performances and its control system are both deterministic, as given inputs will always results in the same outputs. On the other hand, the atmospheric turbulence and the errors introduced by the onboard sensors are stochastic and continuously varying.}. This duality also occurs in the high fidelity simulation, in which the deterministic or stochastic nature of each process is present in the simulation module that models it, hence increasing the realism of the results. As such, a single simulated flight is not sufficient to evaluate the performance of a given GNSS-Denied navigation system, as the randomness implicit to all stochastic processes may result in conditions that are favorable or unfavorable for the algorithm being evaluated, leading to results that are too optimistic or pessimistic.

The most rigorous way of testing any proposed navigation algorithm is to evaluate it by means of a Monte Carlo simulation, in which a sufficiently high number of executions or runs (\nm{\nEX}) of a certain scenario are performed, and the results are aggregated to obtain meaningful and realistic metrics that describe the algorithm behavior. It is also necessary to ensure that the stochastic parameters, in addition to randomly changing between different Monte Carlo executions, are also repeatable if a given run needs to be executed a second time. 

Given the high number of stochastic parameters influencing each flight, it is necessary to employ a high number \nm{\nEX} of executions or runs in each Monte Carlo simulation, with 100 considered as the default. The process begins with the initialization of a discrete uniform distribution with a given seed (any value is valid, so 1 was employed by the author), which generates pseudo random integers in which each possible value has an equal likelihood of being produced. This distribution is realized \nm{\nEX} times to generate the trajectory seeds \nm{\seed, \, j = 1 \dots \nEX}. The stored trajectory seeds \nm{\seed} become the initialization seeds for each of the flight simulation executions, so this step does not need to be repeated.

Every time the Monte Carlo simulation needs to generate a trajectory, it is initialized with the corresponding trajectory seed \nm{\seed}. As this seed is the only input required for all the stochastic processes within the simulation, the results of a given run can always be repeated by employing the same seed. At the beginning of each execution, the trajectory seed \nm{\seed} is employed to initialize a second discrete uniform distribution, which is realized seventeen times (always in the same order to maintain repeatability) to generate the seeds employed by the different simulation modules. These seeds are listed in table \ref{tab:seeds}, and their purpose is explained when applicable later in this article.
\begin{center}
\begin{tabular}{ll}
	\hline
	\textbf{Seed} & \textbf{Purpose} \\
	\hline
	\nm{\seedACC, \seedGYR, \seedMAG}  		& Accelerometers, gyroscopes, and magnetometers \\
	\nm{\seedOSP, \seedOAT, \seedGNSS} 		& Barometer, thermometer, and GNSS receiver \\
	\nm{\seedTAS, \seedAOA, \seedAOS}  		& Pitot tube and air vanes \\	
	\nm{\seedPLAT}					   		& IMU position and attitude \\
	\nm{\seedCAM}                      		& Digital camera \\
	\nm{\seedWIND, \seedWEATHER, \seedTURB} & Wind, weather, and turbulence \\ 	
	\nm{\seedMISSION}                       & Mission guidance objectives \\
	\nm{\seedGEO}                           & Gravity and magnetic field realism \\
	\nm{\seedALIGN}                         & Fine alignment results \\
	\hline
\end{tabular}
\end{center}
\captionof{table}{Seeds required to initialize the different simulation stochastic processes} \label{tab:seeds}


\section{High Fidelity Simulation}\label{sec:HighFidelitySimulation}

This article presents a stochastic high fidelity simulation of the flight in turbulent conditions and varying weather of an autonomous fixed wing low SWaP vehicle, with the purpose of testing the capabilities of different inertial, visual, and visual inertial navigation algorithms in GNSS-Denied conditions. The simulation open source \nm{\CC} code is available in \cite{Gallo2020_simulation}. It is intended to serve as a test bench for the development and evaluation of different GNSS-Denied navigation techniques. For that purpose, the various components of the simulation have been modeled with as few simplifications as possible to increase the realism of the results under two different scenarios, one representing the challenges involved in aborting the mission and heading towards a remote recovery location while experiencing varying weather, and the second modeling the continuation of the mission based on a series of closely spaced bearing changes.
\begin{figure}[h]
\centering
\begin{tikzpicture}[auto, node distance=2cm,>=latex']
	\node [coordinate](x0input) {};
	\node [coordinate, below of=x0input, node distance=3.0cm] (deltaCNTRinput){};
	\node [blockyellow, right of=x0input, minimum width=1.8cm, node distance=3.0cm, align=center, minimum height=1.25cm] (FLIGHT) {\texttt{FLIGHT}};
	\node [coordinate, below of=FLIGHT, node distance=1.75cm] (midblocks){};
	\node [blockyellow, right of=midblocks, minimum width=1.8cm, node distance=1.25 cm, align=center, minimum height=1.25cm] (EARTH) {\texttt{EARTH}};
	\node [blockyellow, left of=midblocks, minimum width=1.8cm, node distance=1.25 cm, align=center, minimum height=1.25cm] (AIRCRAFT) {\texttt{AIRCRAFT}};
	\node [coordinate, right of=FLIGHT, node distance=2.8cm] (crosspoint1){};
	\node [coordinate, below of=crosspoint1, node distance=2.7cm] (crosspoint2){};
	\node [coordinate, left of=crosspoint2, node distance=5.5cm] (crosspoint3){};
	\node [blockgreen, right of=FLIGHT, minimum width=2.6cm, node distance=5.0cm, align=center, minimum height=1.25cm] (SENSORS) {\texttt{SENSORS}};
	\node [blockgreen, right of=SENSORS, minimum width=2.6cm, node distance=5.0cm, align=center, minimum height=1.25cm] (NAVIGATION) {\texttt{NAVIGATION}};
	\node [coordinate, right of=NAVIGATION, node distance=2.6cm] (x0estinput){};
	\node [coordinate, below of=NAVIGATION, node distance=1.5cm](tmp1) {};
	\node [blockgreen, below of=SENSORS, minimum width=2.6cm, node distance=3.0cm, align=center, minimum height=1.25cm] (CONTROL) {\texttt{CONTROL}};
	\node [blockgreen, below of=NAVIGATION, minimum width=2.6cm, node distance=3.0cm, align=center, minimum height=1.25cm] (GUIDANCE) {\texttt{GUIDANCE}};
	\node [coordinate, right of=GUIDANCE, node distance=2.6cm](xrefinput) {};

	\draw [->] ($(EARTH.north)-(0.50cm,0cm)$) -- ($(FLIGHT.south)+(0.75cm,0cm)$);
	\draw [->] (EARTH.west) -- (AIRCRAFT.east);
	\draw [->] ($(AIRCRAFT.north)+(0.50cm,0cm)$) -- ($(FLIGHT.south)-(0.75cm,0cm)$);
	\draw [->] (x0input) -- node[pos=0.4] {\nm{\xveczero}} (FLIGHT.west);
	\draw [->] (x0estinput) -- node[pos=0.4] {\nm{\xvecestzero}} (NAVIGATION.east);
	\draw [->] (FLIGHT.east) -- node[pos=0.5] {\nm{\xvec = \xTRUTH}} (SENSORS.west);
	\draw [->] (SENSORS.east) -- node[pos=0.5] {\nm{\xvectilde = \xSENSED}} (NAVIGATION.west);
	\filldraw [black] (tmp1) circle [radius=1pt];
	\draw [->] (NAVIGATION.south) -- node[pos=0.95] {\nm{\xvecest = \xEST}} (tmp1) -| (CONTROL.north);
	\draw [->] (tmp1) -- (GUIDANCE.north);
	\draw [->] (xrefinput) -- node[pos=0.4] {\nm{\xREF}} (GUIDANCE.east);
	\draw [->] (GUIDANCE.west) -- node[pos=0.5] {\nm{\deltaTARGET}} (CONTROL.east);
	\draw [->] (CONTROL.west) -| node[pos=0.25] {\nm{\deltaCNTR}} (deltaCNTRinput) |- ($(AIRCRAFT.west)+(0cm,0.4cm)$);
	\filldraw [black] (crosspoint1) circle [radius=1pt];
	\draw [->] (crosspoint1) |- (EARTH.east);
	\draw [->] (crosspoint1) -- (crosspoint2) -- (crosspoint3) |- ($(AIRCRAFT.west)-(0cm,0.4cm)$);
\end{tikzpicture}
\caption{Components of the high fidelity simulation}
\label{fig:flow_diagram}
\end{figure}
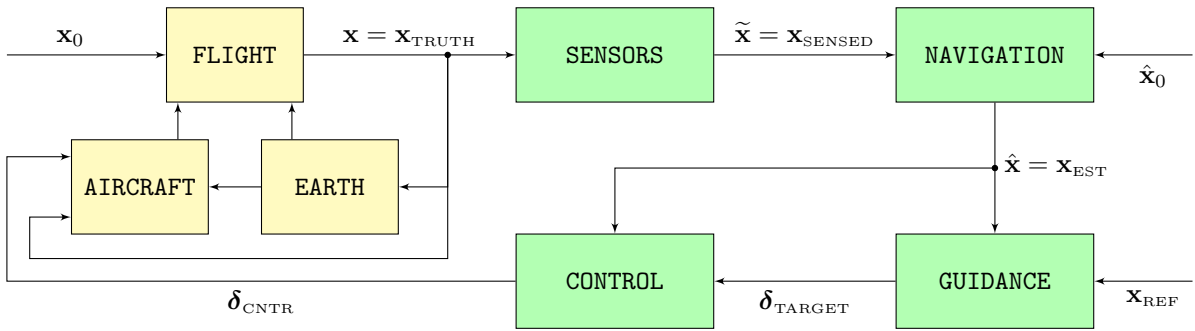

It is necessary to remark that the focus is on low SWaP autonomous UAVs, ruling out the use of high quality sensors, which in general are bigger and have more weight, heavier platforms with more inertia against atmospheric turbulence, as well as the assistance of any kind of communications between the platform and the ground. Platforms that generate their lift by means of rotating blades (helicopters and multirotors) are also excluded from the simulation. Vertical Take Off and Landing (VTOL) as well as Short Take Off and Landing (STOL) vehicles, which use their rotors to generate lift when the airspeed is low, but otherwise are capable of flying like conventional fixed wing aircraft, can be included but only when behaving like fixed wing platforms. Although the modeled platform employs a piston engine to power the propeller, there would be no significant difference if it were replaced by an electric motor.

The objective is to develop and implement a sufficiently realistic simulation so as to enable the implementation and testing of new navigation algorithms to verify if they result in improvements when operating in GNSS-Denied conditions, and if they do, to accurately evaluate the benefits. The simulation models the influence on the resulting aircraft trajectory of many different factors, such as the atmospheric conditions, the wind field, the air turbulence, the aircraft aerodynamic and propulsive performances, its onboard sensors and their error sources, the guidance objectives that make up the mission, the control system that moves the throttle and the aerodynamic controls so the trajectory conforms to the guidance objectives, and the navigation system that processes the data obtained by the sensors and feeds the control system. Figure \ref{fig:flow_diagram} shows the different components of the simulation.

The simulation consists on two distinct processes. The first, represented by the yellow blocks on the left of figure \ref{fig:flow_diagram}, focuses on the physics of flight and the interaction between the aircraft and its surroundings that results in the actual or real aircraft trajectory \nm{\xvec = \xTRUTH}; the second, represented by the green blocks on the right, contains the aircraft systems in charge of ensuring that the resulting trajectory adheres as much as possible to the mission objectives. It includes the different sensors whose output comprise the sensed trajectory \nm{\xvectilde = \xSENSED}, the navigation system in charge of filtering it to obtain the estimated trajectory \nm{\xvecest = \xEST}, the guidance system that converts the reference objectives \nm{\xREF} into the control targets \nm{\deltaTARGET}, and the control system that adjusts the position of the throttle and aerodynamic control surfaces \nm{\deltaCNTR} so the estimated trajectory \nm{\xvecest} is as close as possible to \nm{\xREF}. As shown in the figure, the two parts of the simulation are not independent. The total error or difference between \nm{\xvec} and \nm{\xREF} is the combination of the navigation system error (difference between \nm{\xvec} and \nm{\xvecest}) and the flight technical error (difference between \nm{\xvecest} and \nm{\xREF}). The different simulation components are described below:


\subsubsection*{Earth}

The Earth model conforms the underlying frame in which the aircraft motion and its interaction with the environment takes place:
\begin{itemize}
    \item The World Geodetic System 1984 (WGS84) \cite{WGS84} ellipsoid is employed for the Earth surface.
    \item The Somigliana ellipsoidal formula provided by \cite{WGS84} is adopted for the gravity acceleration \nm{\gc} as an approximation to the combination of the Earth Gravitational Model 1996 (EGM96) \cite{EGM96} gravitation and the Earth centrifugal acceleration.
    \item An spherical Earth of radius \nm{\RE} \cite{SMITHSONIAN} is employed to obtain the ratio between geodetic and geopotential altitudes (h and H) \cite{ISA}.
    \item The wind is modeled as the combination of its low and high frequency components (wind field and turbulence, respectively). The wind field provides the horizontal wind (intensity and direction) as a function of time and geodetic coordinates \nm{\xEgdt} (longitude \nm{\lambda}, latitude \nm{\varphi}, and altitude h) and considers zero vertical wind. The Dryden model \cite{Turbulence1980} is employed for the turbulence or high frequency wind.
    \item A generalization of the International Civil Aviation Organization (ICAO) Standard Atmosphere (ISA) \cite{ISA} is used to model the atmosphere (relationships between pressure p, temperature T, pressure altitude \nm{\Hp}, and geopotential altitude H). Called ICAO Non Standard Atmosphere or INSA \cite{INSA}, it relies on the same hypotheses as ISA but extends it with two parameters, the temperature and pressure offsets (\nm{\DeltaT} and \nm{\Deltap}), which are zero for ISA. In this way, the temperature in standard mean sea level conditions (\nm{\Hp = 0}) is \nm{\Tzero + \DeltaT} instead of \nm{\Tzero} for ISA, while the pressure in mean sea level conditions (\nm{H = 0}) is \nm{\pzero + \Deltap} instead of \nm{\pzero} for ISA.
    \item The weather model provides the temperature and pressure offsets as functions of time and horizontal position (longitude and latitude).
    \item The World Magnetic Model (WMM) is employed for the Earth magnetic field \cite{WMM}.
	\item In order to represent the EGM96 and WMM model errors as well as local effects that result in the onboard models never being fully equal to the real magnetism and gravity experienced by the aircraft, the magnetic and gravity models employed when simulating the aircraft motion are slightly different than those employed by the navigation filter, with the differences stochastically taken from \cite{WMM, Madden2006}.
\end{itemize}

The ellipsoid, gravity field, INSA atmospheric relationships, and Earth magnetism are all deterministic and do not vary. The atmospheric turbulence encountered by the aircraft during its flight and the local variations between the real gravity and magnetic fields experienced by the aircraft with the models employed by the navigation system are stochastic, being randomly generated for each execution \nm{\seed} based on their respective seeds \nm{\seedTURB} and \nm{\seedGEO}. Finally, although the weather and wind fields consist of user provided inputs, the two scenarios described below define them according to stochastic parameters randomly generated based on the \nm{\seedWEATHER} and \nm{\seedWIND} seeds.


\subsubsection*{Aircraft}

The aircraft model provides the aerodynamic and propulsive forces and moments generated by the platform, together with the fuel consumption, the center of gravity position, and the inertia matrix. The modeled platform is a high wing aircraft (maximum mass \nm{19.715 \lrsb{kg}}, length \nm{2.32 \lrsb{m}}, wing span \nm{2.68 \lrsb{m}}) equipped with a \nm{\wedge} tail, a two stroke atmospheric gasoline engine with a maximum power of \nm{4.18 \lrsb{kW}}, and a two blade fixed pitch pusher propeller with a diameter of \nm{0.51 \lrsb{m}}.
\begin{figure}[!h]
	\centering \includegraphics[width=8cm]{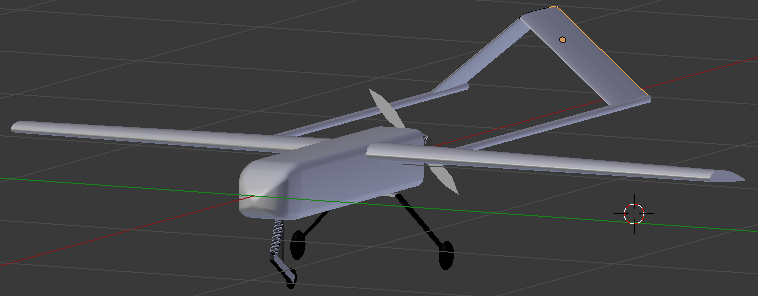}
    \caption{Aircraft overall view}\label{fig:aircraft}
\end{figure}
\begin{itemize}
    \item Mass and inertia analysis is employed to provide the quasi stationary relationship between the inertia matrix and the center of gravity position with the fuel load.
    \item Athena Vortex Lattice \cite{AVL} is used to obtain the dependencies of the platform aerodynamic force and moment coefficients with the angles of attack \nm{\alpha} and sideslip \nm{\beta}, the position of the control surfaces \nm{\deltaCNTR} (elevator, ailerons, and rudder), and the platform angular velocity components \cite{Eshelby2000}. The results have been verified for longitudinal and lateral stability \cite{Etkin1972}. Incompressible air flow is employed to compute the aerodynamic forces and moments.
    \item The propulsive force (thrust) and moment (propeller torque) are obtained by combining a model that provides the engine power and fuel consumption as functions of the throttle position and atmospheric pressure and temperature \cite{Heywood1988}, with a propeller model that provides its thrust and power coefficients based on the advance ratio \cite{Hill1992}.
\end{itemize}

The aircraft performances are deterministic and do not vary for the different executions of the Monte Carlo simulation.


\subsubsection*{Flight}

Given the initial conditions \nm{\xveczero} and the variation with time of the throttle and control surfaces \nm{\deltaCNTR\lrp{t}}, the actual or real trajectory \nm{\xvec\lrp{t}} is the outcome of the interaction between the aircraft and the environment that surrounds it. Considering as inertial a reference frame centered at the Sun with axes fixed with respect to the stars, it is possible to establish a system of first order differential equations composed by the kinematic, force, rotation, moment, and mass equations, which is based on a state vector comprised by the aircraft geodetic coordinates \nm{\xEgdt}, absolute velocity viewed in the body frame \nm{\vB}, attitude from NED to body represented by its unit quaternion \nm{\qNB}, inertial angular velocity viewed in body \nm{\wIBB}, and mass m \cite{Etkin1972}. Note that the Earth and aircraft performance algebraic equations are indirectly included in the system of differential equations, as shown in figure \ref{fig:flow_diagram}.
\begin{eqnarray}
\nm{\xvec\lrp{t}} & = & \nm{\lrsb{\xEgdt\lrp{t}, \, \vB\lrp{t}, \, \qNB\lrp{t}, \, \wIBB\lrp{t}, \, m\lrp{t}}^T}\label{eq:flight_state_vector} \\
\nm{\xvecdot\lrp{t}} & = & \nm{\vec f \big(\xvec\lrp{t}, \, t, \, \deltaCNTR\lrp{t}\big)}\label{eq:flight_differentials} \\
\nm{\xvec\lrp{0}} & = & \nm{\xveczero}\label{eq:flihgt_initial}
\end{eqnarray}

A fourth order Runge-Kutta scheme at a frequency of \nm{500 \lrsb{hz}} is employed to integrate it and obtain the actual trajectory \nm{\xvec\lrp{t_t} = \xTRUTH\lrp{t_t} = \xvec\lrp{t \cdot \DeltatTRUTH}}. A standard Runge-Kutta integration scheme implies considering the unit quaternions as 4-vectors that belong to \nm{\mathbb{R}^4} instead of \nm{\mathbb{SO}(3)} and operating with the addition and scalar multiplication of the \nm{\mathbb{R}^4} vector space; the propagated 4-vectors hence do not belong to \nm{\mathbb{SO}\lrp{3}}, this is, are not unit quaternions, and must be normalized after each step to eliminate the deviation from the three dimensional manifold \nm{\mathbb{SO}\lrp{3}} incurred by the \nm{\mathbb{R}^4} propagation. 

Although at \nm{500 \lrsb{hz}} the errors are relatively small, it is more rigorous and precise to integrate the unit quaternion respecting the constraints of its Lie group, this is, ensuring that \nm{\qNB} propagates without deviating from the \nm{\mathbb{SO}(3)} manifold. To do so, the fourth order Runge-Kutta scheme can be modified so its relies on the \nm{\mathbb{SO}(3)} \nm{\oplus} operator and the relationship between the \nm{\qNB} unit quaternion and the angular velocity \nm{\wNBB} belonging to its local tangent space \nm{\mathfrak{so}(3)} \cite{Sola2017}. Note that \nm{\wNBB} can easily be obtained from \nm{\wIBB} and the rest of the state vector components.

There are no stochastic parameters influencing the integration of the equations of motion, which is a fully deterministic process.


\subsubsection*{Sensors}

The onboard sensors measure various aspects of the actual trajectory and provide these measurements to the navigation system. It is assumed that all sensors are attached to the aircraft structure in a strapdown configuration, provide instantaneous measurements, and are time synchronized with each other. With the exception of the GNSS receiver that operates at \nm{1 \lrsb{hz}} and the digital camera that works at \nm{10 \lrsb{hz}}, all other sensors obtain measurements at a frequency of \nm{100 \lrsb{hz}}. The sensed trajectory \nm{\xvectilde = \xSENSED} is composed by the specific force \nm{\fIBBtilde} measured by the accelerometers, the inertial angular velocity \nm{\wIBBtilde} provided by the gyroscopes, the magnetic field \nm{\BBtilde} measured by the magnetometers, the geodetic coordinates \nm{\widetilde{\xvec}_{\sss GDT}} and ground velocity \nm{\vNtilde} provided by the GNSS receiver (when operative), the atmospheric pressure \nm{\widetilde{p}} and temperature \nm{\widetilde{T}} provided by the barometer and thermometer, the true airspeed \nm{\vtastilde} measured by the Pitot tube, the angles of attack \nm{\widetilde{\alpha}} and sideslip \nm{\widetilde{\beta}} provided by the air vanes, and the digital image of the ground \nm{\vec I} obtained by the camera:
\neweq{\xvectilde\lrp{t_s} = \xvectilde\lrp{s \cdot \DeltatSENSED} = \lrsb{\fIBBtilde, \, \wIBBtilde, \, \BBtilde, \, \widetilde{\xvec}_{\sss GDT}, \, \vNtilde, \, \widetilde{p}, \, \widetilde{T}, \, \vtastilde, \, \widetilde{\alpha}, \, \widetilde{\beta}, \, \vec I}^T}{eq:sensors}

This section provides a summary of the modeling of the error sources of the various sensors, which are described in detail in \cite{SENSORS}. Inertial sensors (accelerometers and gyroscopes) are modeled taking into consideration fixed error sources (scale factor, cross coupling, lever arm, IMU body attitude), run-to-run bias offset, and in-run bias drift and system noise \cite{Farrell2008, Groves2008, Chatfield1997}. The bias and system noise models are based on \cite{Crassidis2006, Woodman2007}. Inertial sensors are assumed to be lab calibrated, which reduces their scale factor and cross coupling errors. 

The simulation considers that the IMU lever arm or position with respect to the aircraft structure is fixed or deterministic. However, its orientation with respect to the body frame is stochastic and hence varies randomly for each execution \nm{\seed} based on \nm{\seedPLAT}. The IMU knowledge about its own position and attitude with respect to body, which are necessary inputs to obtain \nm{\fIBBtilde} and \nm{\wIBBtilde}, is not perfect, and the errors are also stochastically controlled by \nm{\seedPLAT}. Standard normal distributions applied to the inertial sensors specifications are initialized with the \nm{\seedACC} and \nm{\seedGYR} seeds and employed to obtain the value of all fixed, run-to-run, and in-run error contributions.

The magnetometers model is simpler than that of the inertial sensors as it does not include bias drift, although in exchange it needs to consider the negative effects of hard and soft iron magnetism. Additionally, their position and attitude with respect to the aircraft structure have no influence in the readings as long as they are mounted away from permanent magnets and electrical equipment. The scale factor, cross coupling, and hard iron magnetism errors can be reduced if the magnetometers are calibrated through swinging once installed on the aircraft. All error sources are stochastically generated by applying standard normal distributions initialized with \nm{\seedMAG} to the magnetometer specifications.

The barometer, thermometer, Pitot tube, and air vanes (measuring the angles of attack and sideslip) are modeled by a combination of system noise and bias offset exclusively. All parameters vary stochastically in each execution of the Monte Carlo simulation, and are controlled by their respective seeds (\nm{\seedOSP, \, \seedOAT, \, \seedTAS, \, \seedAOA, \, \seedAOS}).

Before the GNSS signals are lost, the navigation system can rely on the position and velocity readings provided by a GNSS receiver. The position error is modeled as the sum of a white noise process plus slow varying ionospheric effects \cite{Kayton1997}, which in turn are modeled as the sum of the bias offset plus a random walk. In the case of the ground velocity error, only system noise is considered. As in previous cases, the simulation errors are obtained by applying standard normal distributions initialized with the \nm{\seedGNSS} seed to the receiver specifications.

The simulation also includes a single onboard camera working at a frequency of 10 [hz] and rigidly attached to the aircraft structure. The simulated image generation process is instantaneous and relies on the shutter speed being sufficiently high that all images are equally sharp. In addition, the camera \texttt{ISO} setting remains constant during the flight, all generated images are noise free, the visible spectrum radiation reaching all patches of the Earth surface remains constant, and in addition the terrain is considered Lambertian, so its appearance at any given time does not vary with the viewing direction. These assumptions imply that a given terrain object is represented with the same luminosity in all images, even as its relative pose (position and attitude) with respect to the camera varies \cite{Soatto2001}. Geometrically, the simulation adopts a perspective projection or pinhole camera model, which in addition is perfectly calibrated and hence shows no distortion \cite{Soatto2001}.

As in the case of the inertial sensors, the simulation considers that the location of the camera onboard the aircraft is deterministic but its orientation is stochastic, as is the navigation system estimation of both the camera position and attitude. All these parameters vary for each execution \nm{\seed} based on the \nm{\seedCAM} seed.

\begin{figure}[h]
\centering
\includegraphics[width=7.5cm]{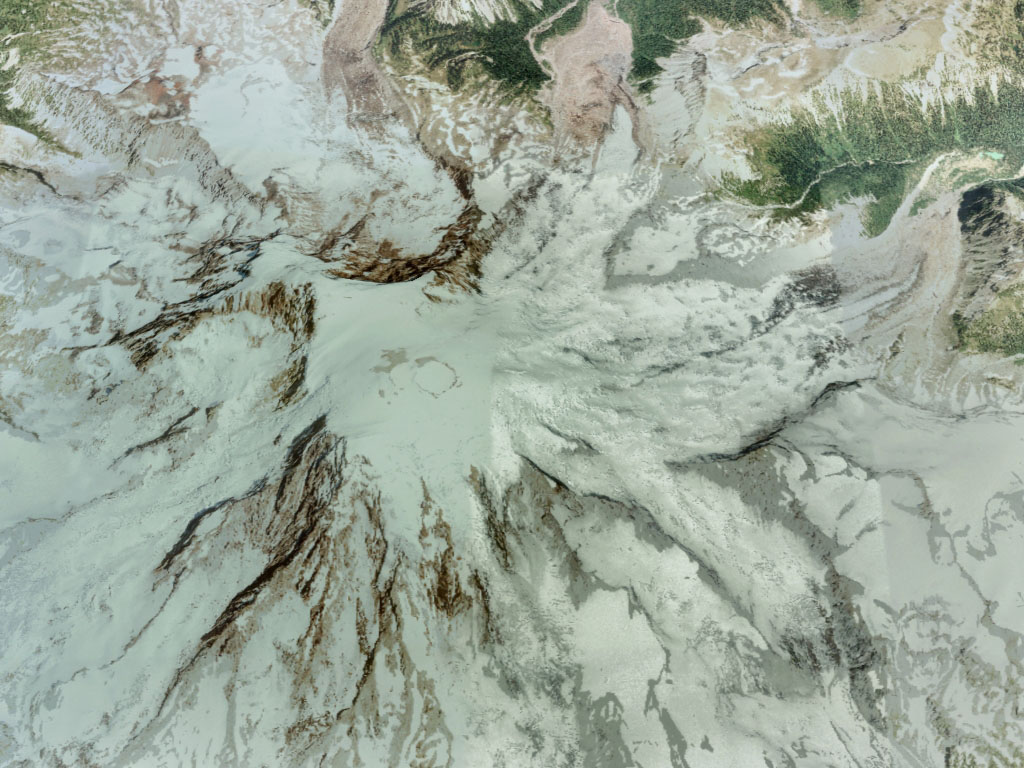}
\hskip 10pt
\includegraphics[width=7.5cm]{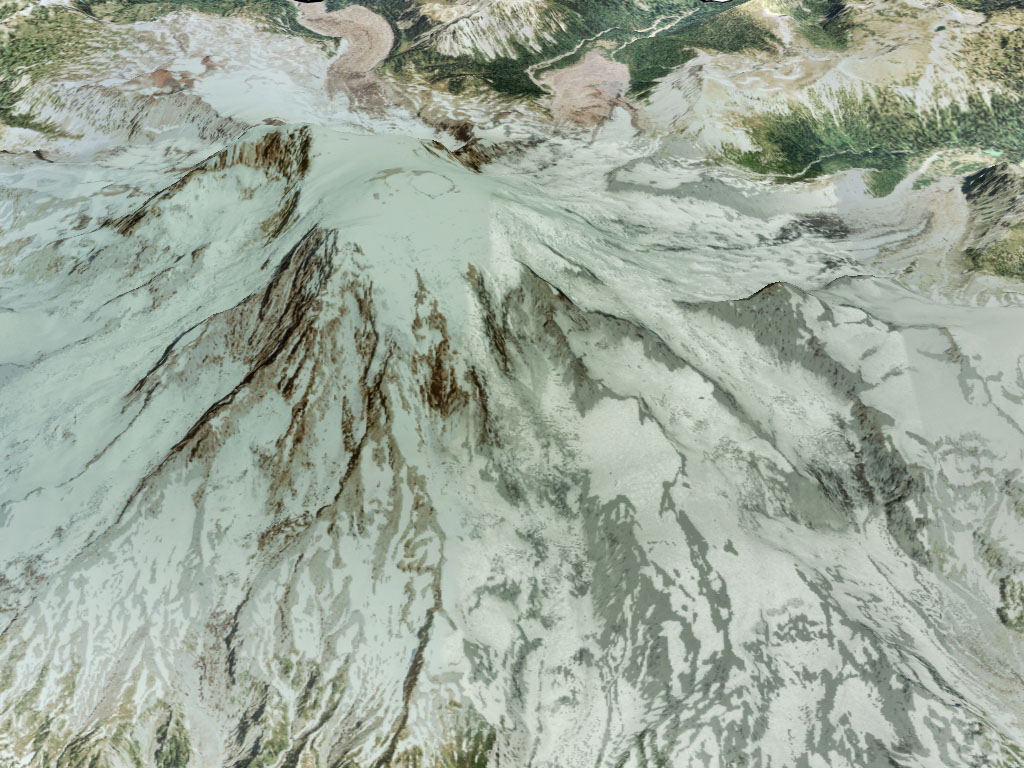}
\caption{Example of \texttt{Earth Viewer} images}
\label{fig:EarthViewer_photo}
\end{figure}

The digital camera differs from all other sensors in that it does not return a sensed variable \nm{\xvectilde} consisting of its real value \nm{\xvec} plus a sensor error \nm{\vec E}, but instead generates a digital image simulating what a real camera would record based on the aircraft position and attitude as given by the actual trajectory \nm{\xvec = \xTRUTH}. When provided with the camera pose at equally time spaced intervals, the simulation is capable of generating images that resemble the view of the Earth surface that the camera would record if located at that particular position and attitude. To do so, it relies on three software libraries:
\begin{itemize}

\item \texttt{OpenSceneGraph} \cite{OpenSceneGraph} is an open source high performance 3D graphics toolkit written in \texttt{C++} and \texttt{OpenGL}, used by application developers in fields such as visual simulation, games, virtual reality, scientific visualization and modeling The library enables the representation of objects in a scene by means of a graph data structure, which allows grouping objects that share some properties to automatically manage rendering properties such as the level of detail necessary to faithfully draw the scene, but without considering the unnecessary detail that slows down the graphics hardware drawing the scene.

\item \texttt{osgEarth} \cite{osgEarth} is a dynamic and scalable 3D Earth surface rendering toolkit that relies on \texttt{OpenSceneGraph}, and is based on publicly available orthoimages of the area flown by the aircraft. Orthoimages consist of aerial or satellite imagery geometrically corrected such that the scale is uniform; they can be used to measure true distances as they are accurate representations of the Earth surface, having been adjusted for topographic relief, lens distortion, and camera tilt. When coupled with a terrain elevation model, \texttt{osgEarth} is capable of generating realistic images based on the camera position as well as its yaw and pitch, but does not accept the camera roll\footnote{This is analogous to stating that \texttt{osgEarth} images are always aligned with the horizon.}.

\item \texttt{Earth Viewer} is a modification to \texttt{osgEarth} implemented by the author so it is also capable of accepting the bank angle of the camera with respect to the \texttt{NED} axes. \texttt{Earth Viewer} is capable of generating realistic Earth images as long as the camera height over the terrain is significantly higher than the vertical relief present in the image. As an example, figure \ref{fig:EarthViewer_photo} shows two different views of a volcano in which the dome of the mountain, having very steep slopes, is properly rendered.
\end{itemize}


\subsubsection*{Guidance and Control}

The guidance and control systems work together so the estimated trajectory \nm{\xvecest = \xEST} resembles the reference objectives \nm{\xREF} as much as possible. In the simulation they both operate at \nm{50 \lrsb{hz}}. The guidance system supplies the control system with the proper targets \nm{\deltaTARGET}, and this in turn adjusts the position of the throttle and aerodynamic control surfaces. As the aircraft has four controls (throttle, elevator, ailerons, and rudder), the guidance system output is at all times composed by four targets and a trigger \nm{\deltaTRG}. The trigger is a boolean expression that is continuously evaluated based on \nm{\xvecest}; when its sign switches, the current control targets are replaced by the next targets in the pipeline.
\begin{eqnarray}
\nm{\xREF} & = & \nm{\lrb{\vec{\delta}_{\sss{TARGET,1}}, \, \vec{\delta}_{\sss{TARGET,2}}, \, \ldots , \vec{\delta}_{\sss{TARGET,n}}}}\label{eq:guidance1} \\
\nm{\deltaTARGET\lrp{t_c}} & = & \nm{\deltaTARGET\lrp{c \cdot \DeltatCNTR} = \lrsb{\deltaTARGETT, \, \deltaTARGETE, \,  \deltaTARGETA, \, \deltaTARGETR, \, \deltaTRG}^T}\label{eq:guidance2} \\
\nm{\deltaCNTR\lrp{t_c}} & = & \nm{\deltaCNTR\lrp{c \cdot \DeltatCNTR} = \lrsb{\deltaT, \ \deltaE, \ \deltaA, \ \deltaR}^T}\label{eq:control}
\end{eqnarray}

The control system is implemented as four primary proportional integral derivative (PID) loops, each associated to a certain variable and control mechanism (true airspeed \nm{\vtas} for the throttle \nm{\deltaT}, body pitch angle \nm{\theta} for the elevator \nm{\deltaE}, body bank angle \nm{\xi} for the ailerons \nm{\deltaA}, and sideslip angle \nm{\beta} for the rudder \nm{\deltaR}). In addition, some secondary PID loops are also employed (pressure altitude \nm{\Hp} and aerodynamic path angle \nm{\gammaTAS} for the elevator, bearing \nm{\chi} for the ailerons), in which the output of the secondary loop acts as the input for the primary one. All loops contain a low pass filter to smooth the derivative inputs \cite{Hagglund2012} and use set point ramping to reduce overshooting.

The guidance and control algorithms have been adjusted to the aircraft performances and are hence deterministic. The reference objectives \nm{\xREF} are however user provided inputs that in the two scenarios described below are randomly generated based on the \nm{\seedMISSION} seed.


\subsubsection*{Navigation}

The mission of the navigation system is to obtain the estimated trajectory \nm{\xvecest = \xEST} based on the sensor outputs \nm{\xvectilde = \xSENSED} and an initial estimation \nm{\xvecestzero}, as shown in figure \ref{fig:flow_diagram}. As the main objective of this simulation is to act as a bench test to analyze the behavior of different inertial and visual navigation systems in GNSS-Denied conditions, no navigation system is included by default as part of the simulation. Follow-on articles will include navigation algorithms proposed by the author as well as the results obtained when applied to Monte Carlo simulations of the two scenarios defined below.

Both inertial and visual navigation systems require an initial estimation of their respective state vectors \nm{\xvecestzero}. While in some occasions these initial estimations can be directly taken from the sensor measurements, other observed trajectory variables are not measured by any sensors, and their initial estimation relies on the pre-flight procedures, and in particular in the outputs of the fine alignment process \cite{Farrell2008, Groves2008, Chatfield1997}. 

When the initial estimation can be directly taken from a given sensor, as is the case with the initial position (provided by the GNSS receiver) required by both inertial and visual systems, no additional randomness is required by the simulation. However, the \nm{\seedALIGN} seed is employed together with a standard normal distribution to estimate the results of the fine alignment process, which  results in initial estimations for the aircraft attitude (required by both inertial and visual systems) as well as the initial errors of the accelerometers, gyroscopes, and magnetometers.


\section{Scenario \#1}\label{sec:Scenario1}

Scenario \#1 has been selected with the objective of adequately representing the challenges faced by an autonomous fixed-wind UAV that suddenly cannot rely on the GNSS signals and hence changes course to reach a predefined recovery location situated at approximately one hour of flight time. In the process, in addition to executing an altitude and airspeed adjustment, the autonomous aircraft faces significant weather and wind field changes that make its GNSS-Denied navigation even more challenging. The following sections describe the different parameters that together define the scenario, with special emphasis on the stochastic ones whose values vary in the different executions or runs of a given Monte-Carlo simulation. 


\subsubsection*{Mission}

The mission, guidance targets, or reference objectives of scenario \#1 consists in the six flight segments graphically shown in figure \ref{fig:Sim_Scenario_Mission_scheme}. The scenario is intended to showcase the ability of an autonomous aircraft to reach a predefined recovery location once it is flying in GNSS-Denied conditions:
\begin{itemize}
\item The first segment starts with the aircraft flying at constant airspeed \nm{\vtasINI}, constant pressure altitude \nm{\HpINI}, constant bearing \nm{\chiINI}, and no sideslip \nm{\beta = 0}. As the navigation filter is employing the GNSS signals, the solution is stable and hence the difference between the observed trajectory \nm{\xEST = \xvecest} and the actual trajectory \nm{\xTRUTH = \xvec} does not increase with time. For this reason, there is no lack in generality in the assumption that the GNSS signal gets lost at \nm{\tGNSS = 100 \, \lrsb{sec}}\footnote{The results would be similar if the aircraft flew with GNSS for a much longer period of time.}. It is also realistic to assume that the aircraft will continue flying in the same way for a short period of time until confirming that it can not recover the GNSS signal, and then change its mission at \nm{\tTURN} by performing a bearing change in an attempt to reach a predefined recovery point. 

\item The turn is executed at the same airspeed \nm{\vtasINI} and pressure altitude \nm{\HpINI} with a body bank angle of \nm{\xiTURN = \pm 10 \, \lrsb{deg}}, with the sign given by the shortest turn necessary to reach the final bearing \nm{\chiEND}, upon which the turn concludes.
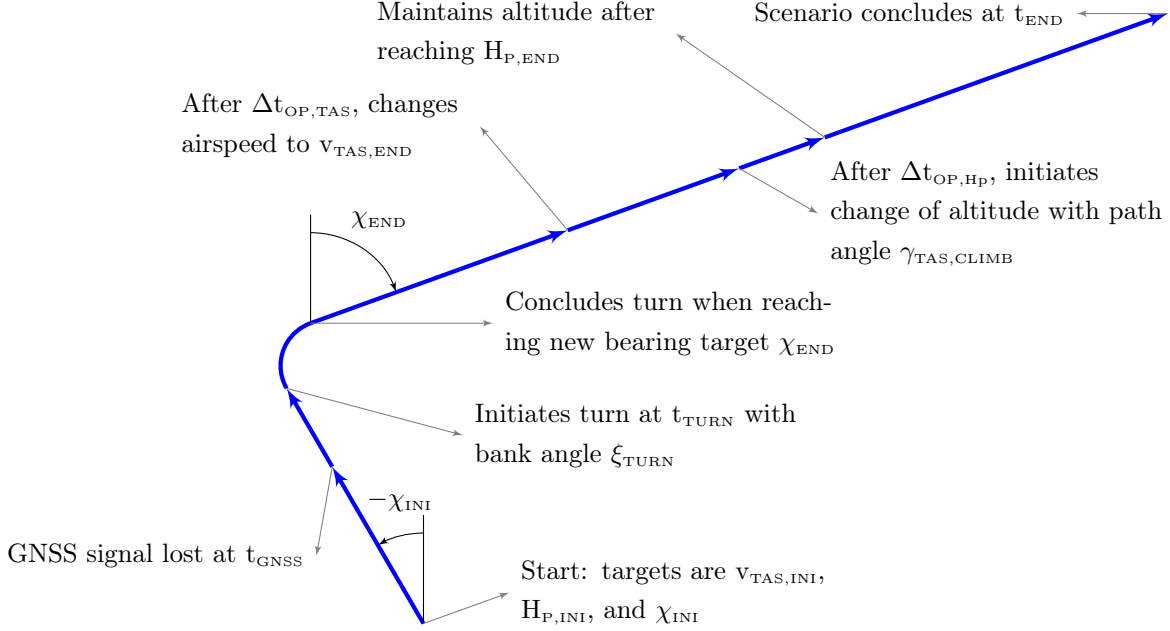
\begin{figure}[h]
\centering
\begin{tikzpicture}[auto,>=latex',scale=1.2]
	\coordinate (center) at (+0.0,+0.0);
	\coordinate (pointC) at ($(center) + (210:0.5cm)$);
	\coordinate (pointD) at ($(center) + (110:0.5cm)$);
	\coordinate (pointB) at ($(pointC) + (-60:1.0cm)$);
	\coordinate (pointA) at ($(pointC) + (-60:3.0cm)$);
	\coordinate (pointE) at ($(pointD) + (20:3.0cm)$);
	\coordinate (pointF) at ($(pointD) + (20:5.0cm)$);
	\coordinate (pointG) at ($(pointD) + (20:6.0cm)$);
	\coordinate (pointH) at ($(pointD) + (20:10.0cm)$);
			
	\draw [blue, name path = segmentAB, ultra thick] [->] (pointA) -- (pointB);
	\draw [blue, name path = segmentBC, ultra thick] [->] (pointB) -- (pointC);
	\draw (pointC) [blue, name path = segmentCD, ultra thick] arc[start angle=210, end angle=110, radius=0.5cm];
	\draw [blue, name path = segmentDE, ultra thick] [->] (pointD) -- (pointE);	
	\draw [blue, name path = segmentEF, ultra thick] [->] (pointE) -- (pointF);	
	\draw [blue, name path = segmentFG, ultra thick] [->] (pointF) -- (pointG);	
	\draw [blue, name path = segmentGH, ultra thick] [->] (pointG) -- (pointH);	
	
	\coordinate (pointAref) at ($(pointA) + (90:1.2cm)$);
	\draw [name path = Aref, thin] [-] (pointA) -- (pointAref);
	\draw ($(pointA) + (90:1.0cm)$) [name path = pointAangle, thin] arc[start angle=90, delta angle=30, radius=1.0cm] [->] node [pos=0.5, above=6pt] {\nm{- \chiINI}};
	
	\coordinate (pointDref) at ($(pointD) + (90:1.2cm)$);
	\draw [name path = Dref, thin] [-] (pointD) -- (pointDref);
	\draw ($(pointD) + (90:1.0cm)$) [name path = pointDangle, thin] arc[start angle=90, delta angle=-70, radius=1.0cm] [->] node [pos=0.7, above=10pt] {\nm{\chiEND}};
	
	\coordinate (pointAA) at ($(pointA) + (+20:1.0cm)$);
	\draw [black!50, thin] [->] (pointA) -- (pointAA) node [black, right=1pt, text width=4.5cm] {Start: targets are \nm{\vtasINI}, \nm{\HpINI}, and \nm{\chiINI}};
	
	\coordinate (pointBB) at ($(pointB) + (+260:1.0cm)$);
	\draw [black!50, thin] [->] (pointB) -- (pointBB) node [black, left=1pt] {GNSS signal lost at \nm{\tGNSS}};
		
	\coordinate (pointCC) at ($(pointC) + (-15:2.0cm)$);
	\draw [black!50, thin] [->] (pointC) -- (pointCC) node [black, right=1pt, text width=5cm] {Initiates turn at \nm{\tTURN} with bank angle \nm{\xiTURN}};
		
	\coordinate (pointDD) at ($(pointD) + (+0:2.0cm)$);
	\draw [black!50, thin] [->] (pointD) -- (pointDD) node [black, right=1pt, text width=4.5cm] {Concludes turn when reaching new bearing target \nm{\chiEND}};
	
	\coordinate (pointEE) at ($(pointE) + (+130:1.5cm)$);
	\draw [black!50, thin] [->] (pointE) -- (pointEE) node [black, left=0pt, text width=3.8cm] {After \nm{\DeltatOPTAS}, changes airspeed to \nm{\vtasEND}};

	\coordinate (pointFF) at ($(pointF) + (+330:1.0cm)$);
	\draw [black!50, thin] [->] (pointF) -- (pointFF) node [black, right=1pt, text width=4.5cm] {After \nm{\DeltatOPHp}, initiates change of altitude with path angle \nm{\gammaTASCLIMB}};
	
	\coordinate (pointGG) at ($(pointG) + (+145:2.0cm)$);
	\draw [black!50, thin] [->] (pointG) -- (pointGG) node [black, left=0pt, text width=3.8cm] {Maintains altitude after reaching \nm{\HpEND}};
	
	\coordinate (pointHH) at ($(pointH) + (+180:1.0cm)$);
	\draw [black!50, thin] [->] (pointH) -- (pointHH) node [black, left=1pt] {Scenario concludes at \nm{\tEND}};
		
\end{tikzpicture}
\caption{Schematic view of the scenario \#1 mission}
\label{fig:Sim_Scenario_Mission_scheme}
\end{figure}

Note that the real objective of the aircraft is to reach a predefined recovery location, not to fly a given bearing \nm{\chiEND}. However, one leads to the other, as the aircraft needs to be able to follow a bearing in order to reach a location; and if it is capable of reaching a location, it means that it can also follow a bearing. From the point of view of setting up the scenario and evaluating the results, it is more convenient for each seed to imply the use of a given bearing \nm{\chiEND} for the same length of time (see below) than to convert each seed to a different longitude and latitude which the aircraft would take different times to reach due to wind variations.

\item Once it captures the bearing that should take it to its intended destination, the aircraft follows it at the same airspeed \nm{\vtasINI} and pressure altitude \nm{\HpINI} for a given period of time \nm{\DeltatOPTAS}.

\item At this time the aircraft changes its airspeed to \nm{\vtasEND} without modifying its altitude \nm{\HpINI} or course \nm{\chiEND}. Note that a change of airspeed is not necessary to reach the intended destination but is included to show that the proposed systems are capable of accelerating or decelerating the aircraft without the use of GNSS signals. These conditions are maintained for a given period of time \nm{\DeltatOPHp}.

\item The aircraft then modifies its pressure altitude without changing its airspeed \nm{\vtasEND} or bearing \nm{\chiEND} to avoid the terrain or to get an improved exposure to radio control signals with which it may be recovered from its destination. As in the previous segment, note that this change in pressure altitude is in theory not necessary but is included to prove that it can be achieved in GNSS-Denied conditions. The maneuver is executed with an aerodynamic path angle of \nm{\gammaTASCLIMB = \pm 2 \ \lrsb{deg}}, with the sign given by whether the final pressure altitude \nm{\HpEND} is higher or lower than the existing one. 

\item The last segment involves the aircraft flying at the existing airspeed, pressure altitude, and bearing until time \nm{\tEND = 3800 \, \lrsb{sec}}. The use of such a long flight time (in the order of one hour) in GNSS-Denied conditions ensures that no instabilities or other negative effects in the navigation system are hidden by not allowing the aircraft to fly long enough in GNSS-Denied conditions.
\end{itemize}

The above description of the guidance targets or reference trajectory can also be expressed in a more formal way according to the previously introduced notation:
\begin{eqnarray}
\nm{\xREF}           & = & \nm{\lrb{\deltaTARGETone, \, \deltaTARGETtwo, \, \deltaTARGETthr, \, \deltaTARGETfou, \, \deltaTARGETfiv, \, \deltaTARGETsix}} \label{eq:Sim_Scenario_Mission_RT} \\
\nm{\deltaTARGETone} & = & \nm{\lrsb{\vtasINI, \, \HpINI, \,  \chiINI, \, \beta = 0, \, \tTURN}^T}\label{eq:Sim_Scenario_Mission_target1} \\
\nm{\deltaTARGETtwo} & = & \nm{\lrsb{\vtasINI, \, \HpINI, \,  \xiTURN, \, \beta = 0, \, \chiEND}^T}\label{eq:Sim_Scenario_Mission_target2} \\
\nm{\deltaTARGETthr} & = & \nm{\lrsb{\vtasINI, \, \HpINI, \,  \chiEND, \, \beta = 0, \, \DeltatOPTAS}^T}\label{eq:Sim_Scenario_Mission_target3} \\
\nm{\deltaTARGETfou} & = & \nm{\lrsb{\vtasEND, \, \HpINI, \,  \chiEND, \, \beta = 0, \, \DeltatOPHp}^T}\label{eq:Sim_Scenario_Mission_target4} \\
\nm{\deltaTARGETfiv} & = & \nm{\lrsb{\vtasEND, \, \gammaTASCLIMB, \,  \chiEND, \, \beta = 0, \, \HpEND}^T}\label{eq:Sim_Scenario_Mission_target5} \\
\nm{\deltaTARGETsix} & = & \nm{\lrsb{\vtasEND, \, \HpEND, \,  \chiEND, \, \beta = 0, \, \tEND}^T}\label{eq:Sim_Scenario_Mission_target6}
\end{eqnarray}

\begin{center}
\begin{tabular}{lcccc}
	\hline
	\textbf{Component} & \textbf{Variable} & \textbf{Definition} & \textbf{Unit} & \textbf{Restrictions} \\
	\hline
	Initial Airspeed          & \nm{\vtasINI}       & \nm{N\lrp{29, 1.5^2}}       & [m/sec] & \nm{24 < \vtasINI < 34} \\
	Initial pressure altitude & \nm{\HpINI}         & \nm{N\lrp{2700, 200^2}}     & [m]     & \\
	Initial bearing           & \nm{\chiINI}        & \nm{U\lrp{-179, 180}}       & [deg]   & \\
	Sideslip angle            & \nm{\beta}          & \nm{0}                      & [deg]   & \\
	GNSS-Denied time		  & \nm{\tGNSS}		    & \nm{100}		   		      & [sec]   & \\
	Turn time				  & \nm{\tTURN}		    & \nm{\tGNSS + N(30, 50^2)}   & [sec]   & \nm{\tTURN > \lrp{\tGNSS + 15}} \\
	Turn bank angle			  & \nm{\xiTURN}	    & \nm{\pm 10}			      & [deg]   & \\
	Final bearing             & \nm{\chiEND}        & \nm{U\lrp{-179, 180}}       & [deg]   & \nm{\lvert \chiEND - \chiINI \rvert > 10} \\
	First time interval       & \nm{\DeltatOPTAS}   & \nm{N\lrp{500, 100^2}}      & [sec]   & \nm{\DeltatOPTAS > 150} \\
	\multirow{2}{*}{Final airspeed} & \multirow{2}{*}{\nm{\vtasEND}} & \multirow{2}{*}{\nm{N\lrp{\vtasINI, 1.5^2}}} & \multirow{2}{*}{[m/sec]} & \nm{24 < \vtasEND < 34} \\
	& & & & \nm{\lvert \vtasEND - \vtasINI \rvert > 0.5} \\	
	\multirow{2}{*}{Second time interval} & \multirow{2}{*}{\nm{\DeltatOPHp}} & \multirow{2}{*}{\nm{N\lrp{500, 100^2}}} & \multirow{2}{*}{[sec]} & \nm{\DeltatOPHp > 150} \\
	&  &  &  & \nm{\DeltatOPTAS + \DeltatOPHp < 2500} \\
	Path angle				  & \nm{\gammaTASCLIMB} & \nm{\pm 2}                  & [deg]   & \\
	Final pressure altitude   & \nm{\HpEND}         & \nm{N\lrp{\HpINI, 300^2}}   & [m]     & \nm{\lvert \HpEND - \HpINI \rvert > 100} \\
	Final time                & \nm{\tEND}          & \nm{3800}					  & [sec]   & \\
	\hline
\end{tabular}
\end{center}
\captionof{table}{Deterministic and stochastic components of scenario \#1 mission} \label{tab:Sim_Scenario_Mission_variables}
 
Table \ref{tab:Sim_Scenario_Mission_variables} shows the definition of the deterministic and stochastic parameters that result in the scenario \#1 mission. Note that it has been developed with the objective of including as much variability as possible, as the only deterministic variables are those that do not result in loss of generality. For each Monte Carlo simulation run \nm{\seed}, the \nm{\seedMISSION} seed is employed to initialize a standard normal distribution N and a discrete uniform distribution U, which are then realized several times each to obtain the different parameters listed in table \ref{tab:Sim_Scenario_Mission_variables}. This ensures that the results are repeatable if the same seed is employed again.
\begin{figure}[h]
\centering
\begin{tikzpicture}
\begin{polaraxis}[
colormap name = bluered,
cycle list={[of colormap]},
width=8.0cm,
rotate=-90,
x dir=reverse,
axis lines*=none,
axis line style = {draw=transparent,line width=0.0001pt},
xticklabel style={anchor=-\tick-90},
xtick={0,30,60,90,120,150,180,210,240,270,300,330},
xticklabels={N,30,60,E,120,150,S,-150,-120,W,-60,-30},
ymin=0,
ymax=2.2,
ytick={0,0.5,1,...,2},
yticklabels=\empty,
axis line style={-stealth},
legend entries={\nm{\chiINI \, \lrsb{deg}}, \nm{\chiEND \, \lrsb{deg}}},
legend style={font=\footnotesize},
legend cell align=right,
] 
\pgfplotstableread{figs/seeds_mission_bearing.txt}\mytable
\addplot+ [scatter, only marks, mark=*,       scatter src=explicit] table [header=false, x index=1,y index=2, meta index=0] {\mytable};
\addplot+ [scatter, only marks, mark=square*, scatter src=explicit] table [header=false, x index=3,y index=4, meta index=0] {\mytable};
\end{polaraxis}	
\end{tikzpicture}
\hskip 1pt
\begin{tikzpicture}
\begin{polaraxis}[
colormap name = bluered,
cycle list={[of colormap]},
width=8.0cm,
rotate=-90,
x dir=reverse,
axis lines*=none,
axis line style = {draw=transparent,line width=0.0001pt},
xticklabel style={anchor=-\tick-90},
xtick={0,30,60,90,120,150,180,210,240,270,300,330},
xticklabels={0,30,60,90,120,150,180,-150,-120,-90,-60,-30},
ymin=0,
ymax=2.2,
ytick={0,0.5,1,...,2},
yticklabels=\empty,
axis line style={-stealth},
legend entries={\nm{\chiEND - \chiINI \, \lrsb{deg}}},
legend style={font=\footnotesize},
legend cell align=right,
] 
\pgfplotstableread{figs/seeds_mission_bearing.txt}\mytable
\addplot+ [scatter, only marks, mark=pentagon*,  scatter src=explicit] table [header=false, x index=5,y index=4, meta index=0] {\mytable};
\addplot [black,dashed] coordinates {(+10,+0.0) (+10,+2.2)};
\addplot [black,dashed] coordinates {(-10,+0.0) (-10,+2.2)};
\end{polaraxis}	
\end{tikzpicture}
\caption{Randomness in scenario \#1 mission initial and final bearings}
\label{fig:Sim_Scenario_Mission_seeds_bearing}
\end{figure}
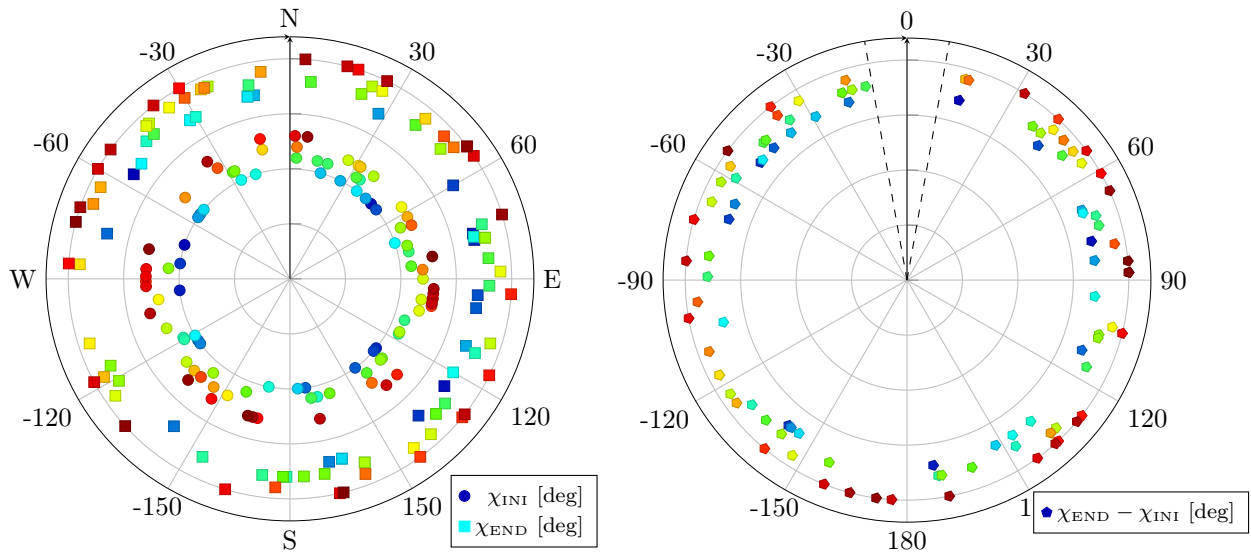

\begin{figure}[h]
\centering
\begin{tikzpicture}
\begin{axis}[
colormap name = bluered,
cycle list={[of colormap]},
width=8.0cm,
xmin=24, xmax=34, xtick={24,26,...,34},
xlabel={\nm{\vtas \, \lrsb{m/\,sec}}},
xmajorgrids,
ymin=+1650, ymax=+3650, ytick={+1650,+2050,...,+3650},
ylabel={\nm{\Hp \, \lrsb{m}}},
ymajorgrids,
axis lines=left,
axis line style={-stealth},
legend entries={\nm{\sss INI}, \nm{\sss END}},
legend style={font=\footnotesize},
legend cell align=left,
] 
\pgfplotstableread{figs/seeds_mission_alt_speed.txt}\mytable
\addplot+ [scatter, only marks, mark=*,       scatter src=explicit] table [header=false, x index=4,y index=1, meta index=0] {\mytable};
\addplot+ [scatter, only marks, mark=square*, scatter src=explicit] table [header=false, x index=5,y index=2, meta index=0] {\mytable};
\end{axis}	
\end{tikzpicture}
\hskip 10pt
\begin{tikzpicture}
\begin{axis}[
colormap name = bluered,
cycle list={[of colormap]},
width=8.0cm,
xmin=-4, xmax=4, xtick={-4,-3,...,4},
xlabel={\nm{\vtasEND - \vtasINI \, \lrsb{m/\,sec}}},
xmajorgrids,
ymin=-750, ymax=+750, ytick={-750,-500,...,750},
ylabel={\nm{\HpEND - \HpINI \, \lrsb{m}}},
ymajorgrids,
axis lines=left,
axis line style={-stealth},
] 
\pgfplotstableread{figs/seeds_mission_alt_speed.txt}\mytable
\addplot+ [scatter, only marks, mark=pentagon*, scatter src=explicit] table [header=false, x index=6,y index=3, meta index=0] {\mytable};
\addplot [black,dashed] coordinates {(-8,+100) (-0.5,+100) (-0.5,+750)};
\addplot [black,dashed] coordinates {(-8,-100) (-0.5,-100) (-0.5,-750)};
\addplot [black,dashed] coordinates {(+8,+100) (+0.5,+100) (+0.5,+750)};
\addplot [black,dashed] coordinates {(+8,-100) (+0.5,-100) (+0.5,-750)};
\end{axis}	
\end{tikzpicture}
\caption{Randomness in scenario \#1 mission initial and final airspeed and pressure altitude}
\label{fig:Sim_Scenario_Mission_seeds_alt_speed}
\end{figure}
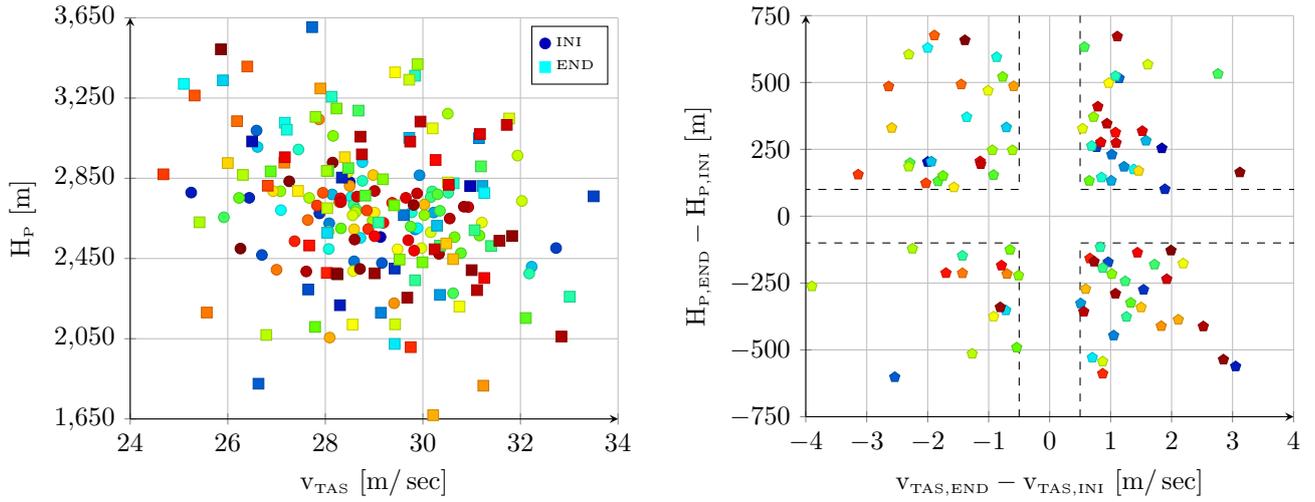

Figures \ref{fig:Sim_Scenario_Mission_seeds_bearing} and \ref{fig:Sim_Scenario_Mission_seeds_alt_speed} are intended to show the wide range of conditions covered by the scenario \#1 mission. The left hand plot of figure \ref{fig:Sim_Scenario_Mission_seeds_bearing} shows the initial bearing \nm{\chiINI} (circles) and final bearing \nm{\chiEND} (squares) for each of the different \nm{\nEX} executions or seeds \nm{\seed} (one color per seed), while the right hand plot shows the bearing change (pentagons) at the turn, with dashed lines representing the restrictions listed in table \ref{tab:Sim_Scenario_Mission_variables}. The simulation hence covers a wide array of cases, from those in which the aircraft barely changes its bearing after loosing the GNSS signals to those in which it turns around and returns with a bearing close to opposite the initial one. Note that the radius has no meaning in this figure, and it is only intended to improve the visualization of the different bearings. 
 
The left hand plot of figure \ref{fig:Sim_Scenario_Mission_seeds_alt_speed} shows the initial (circle) and final (square) pressure altitude and airspeed of the scenario \#1 mission for each of the different \nm{\nEX} executions, while the right hand plot shows  the changes (pentagons) included in the maneuvers described above, with dashed lines representing the restrictions listed in table \ref{tab:Sim_Scenario_Mission_variables}. As in the previous case, the simulation includes cases with small changes in altitude and airspeed, and others with very significant variations. Note that the default airspeed value of \nm{29 \, \lrsb{m/sec}} shown in table \ref{tab:Sim_Scenario_Mission_variables}, as well as the low and high speed limits of \nm{24} and \nm{34 \, \lrsb{m/sec}}, have been selected according to the aircraft performances, while the default pressure altitude of \nm{2700 \, \lrsb{m}} is a compromise between the thrust force diminution with altitude experienced by piston engine aircraft and the need to fly at a sufficiently elevated height over the terrain to ensure that the images generated by the \texttt{Earth Viewer} tool are sharp.
\begin{figure}[h]
\centering
\begin{tikzpicture}
\begin{polaraxis}[
cycle list={{red,no markers,very thick},
           {blue,no markers,very thick},
		   {orange!50!yellow,no markers,very thick},
		   {violet,no markers,very thick},
		   {green,no markers,very thick},
		   {magenta,no markers,very thick},
		   {olive,no markers,very thick}},
width=10.0cm,
rotate=-90,
x dir=reverse,
axis lines*=none,
axis line style = {draw=transparent,line width=0.0001pt},
xticklabel style={anchor=-\tick-90},
xtick={0,30,60,90,120,150,180,210,240,270,300,330},
xticklabels={N,30,60,E,120,150,S,-150,-120,W,-60,-30},
ymin=0,
ymax=160,
ytick={0,25,...,150},
ylabel=km,
axis line style={-stealth},
] 
\pgfplotstableread{figs/error_filter_pos_hor_m.txt}\mytable
\addplot table [header=false, x index= 1,y index= 0] {\mytable};
\addplot table [header=false, x index= 5,y index= 4] {\mytable};
\addplot table [header=false, x index= 9,y index= 8] {\mytable};
\addplot table [header=false, x index=13,y index=12] {\mytable};
\addplot table [header=false, x index=17,y index=16] {\mytable}; 

\addplot table [header=false, x index=21,y index=20] {\mytable};
\addplot table [header=false, x index=25,y index=24] {\mytable};
\addplot table [header=false, x index=29,y index=28] {\mytable};
\addplot table [header=false, x index=33,y index=32] {\mytable};
\addplot table [header=false, x index=37,y index=36] {\mytable}; 

\addplot table [header=false, x index=41,y index=40] {\mytable};
\addplot table [header=false, x index=45,y index=44] {\mytable};
\addplot table [header=false, x index=49,y index=48] {\mytable};
\addplot table [header=false, x index=53,y index=52] {\mytable};
\addplot table [header=false, x index=57,y index=56] {\mytable};

\addplot table [header=false, x index=61,y index=60] {\mytable};
\addplot table [header=false, x index=65,y index=64] {\mytable};
\addplot table [header=false, x index=69,y index=68] {\mytable};
\addplot table [header=false, x index=73,y index=72] {\mytable};
\addplot table [header=false, x index=77,y index=76] {\mytable}; 

\addplot table [header=false, x index=81,y index=80] {\mytable};	
\addplot table [header=false, x index=85,y index=84] {\mytable};	
\addplot table [header=false, x index=89,y index=88] {\mytable};
\addplot table [header=false, x index=93,y index=92] {\mytable};	
\addplot table [header=false, x index=97,y index=96] {\mytable}; 

\addplot table [header=false, x index=101,y index=100] {\mytable};	
\addplot table [header=false, x index=105,y index=104] {\mytable};	
\addplot table [header=false, x index=109,y index=108] {\mytable};	
\addplot table [header=false, x index=113,y index=112] {\mytable};	
\addplot table [header=false, x index=117,y index=116] {\mytable};	 

\addplot table [header=false, x index=121,y index=120] {\mytable};	
\addplot table [header=false, x index=125,y index=124] {\mytable};	
\addplot table [header=false, x index=129,y index=128] {\mytable};	
\addplot table [header=false, x index=133,y index=132] {\mytable};
\addplot table [header=false, x index=137,y index=136] {\mytable};	 

\addplot table [header=false, x index=141,y index=140] {\mytable};	
\addplot table [header=false, x index=145,y index=144] {\mytable};	
\addplot table [header=false, x index=149,y index=148] {\mytable};	
\addplot table [header=false, x index=153,y index=152] {\mytable};	
\addplot table [header=false, x index=157,y index=156] {\mytable};	 

\addplot table [header=false, x index=161,y index=160] {\mytable};	
\addplot table [header=false, x index=165,y index=164] {\mytable};	
\addplot table [header=false, x index=169,y index=168] {\mytable};	
\addplot table [header=false, x index=173,y index=172] {\mytable};	
\addplot table [header=false, x index=177,y index=176] {\mytable};	 

\addplot table [header=false, x index=181,y index=180] {\mytable};	
\addplot table [header=false, x index=185,y index=184] {\mytable};	
\addplot table [header=false, x index=189,y index=188] {\mytable};	
\addplot table [header=false, x index=193,y index=192] {\mytable};	
\addplot table [header=false, x index=197,y index=196] {\mytable};	 

\addplot table [header=false, x index=201,y index=200] {\mytable};	
\addplot table [header=false, x index=205,y index=204] {\mytable};	
\addplot table [header=false, x index=209,y index=208] {\mytable};	
\addplot table [header=false, x index=213,y index=212] {\mytable};	
\addplot table [header=false, x index=217,y index=216] {\mytable};	

\addplot table [header=false, x index=221,y index=220] {\mytable};	
\addplot table [header=false, x index=225,y index=224] {\mytable};	
\addplot table [header=false, x index=229,y index=228] {\mytable};	
\addplot table [header=false, x index=233,y index=232] {\mytable};	
\addplot table [header=false, x index=237,y index=236] {\mytable};	 

\addplot table [header=false, x index=241,y index=240] {\mytable};
\addplot table [header=false, x index=245,y index=244] {\mytable};	
\addplot table [header=false, x index=249,y index=248] {\mytable};	
\addplot table [header=false, x index=253,y index=252] {\mytable};	
\addplot table [header=false, x index=257,y index=256] {\mytable};	 

\addplot table [header=false, x index=261,y index=260] {\mytable};	
\addplot table [header=false, x index=265,y index=264] {\mytable};
\addplot table [header=false, x index=269,y index=268] {\mytable};	
\addplot table [header=false, x index=273,y index=272] {\mytable};
\addplot table [header=false, x index=277,y index=276] {\mytable};	 

\addplot table [header=false, x index=281,y index=280] {\mytable};	
\addplot table [header=false, x index=285,y index=284] {\mytable};	
\addplot table [header=false, x index=289,y index=288] {\mytable};	
\addplot table [header=false, x index=293,y index=292] {\mytable};	
\addplot table [header=false, x index=297,y index=296] {\mytable}; 

\addplot table [header=false, x index=301,y index=300] {\mytable};
\addplot table [header=false, x index=305,y index=304] {\mytable};	
\addplot table [header=false, x index=309,y index=308] {\mytable};	
\addplot table [header=false, x index=313,y index=312] {\mytable};	
\addplot table [header=false, x index=317,y index=316] {\mytable}; 

\addplot table [header=false, x index=321,y index=320] {\mytable};
\addplot table [header=false, x index=325,y index=324] {\mytable};	
\addplot table [header=false, x index=329,y index=328] {\mytable};	
\addplot table [header=false, x index=333,y index=332] {\mytable};	
\addplot table [header=false, x index=337,y index=336] {\mytable};	 

\addplot table [header=false, x index=341,y index=340] {\mytable};	
\addplot table [header=false, x index=345,y index=344] {\mytable};	
\addplot table [header=false, x index=349,y index=348] {\mytable};	
\addplot table [header=false, x index=353,y index=352] {\mytable};	
\addplot table [header=false, x index=357,y index=356] {\mytable};	

\addplot table [header=false, x index=361,y index=360] {\mytable};	
\addplot table [header=false, x index=365,y index=364] {\mytable};	
\addplot table [header=false, x index=369,y index=368] {\mytable};
\addplot table [header=false, x index=373,y index=372] {\mytable};	
\addplot table [header=false, x index=377,y index=376] {\mytable};	 

\addplot table [header=false, x index=381,y index=380] {\mytable};	
\addplot table [header=false, x index=385,y index=384] {\mytable};	
\addplot table [header=false, x index=389,y index=388] {\mytable};	
\addplot table [header=false, x index=393,y index=392] {\mytable};	
\addplot table [header=false, x index=397,y index=396] {\mytable};	 

\end{polaraxis}	
\end{tikzpicture}
\caption{GNSS-Based horizontal tracks for scenario \#1 seeds \nm{\Upsilon_{1}} through \nm{\Upsilon_{100}}}
\label{fig:Sim_NSE_Results_xhor_radar}
\end{figure}
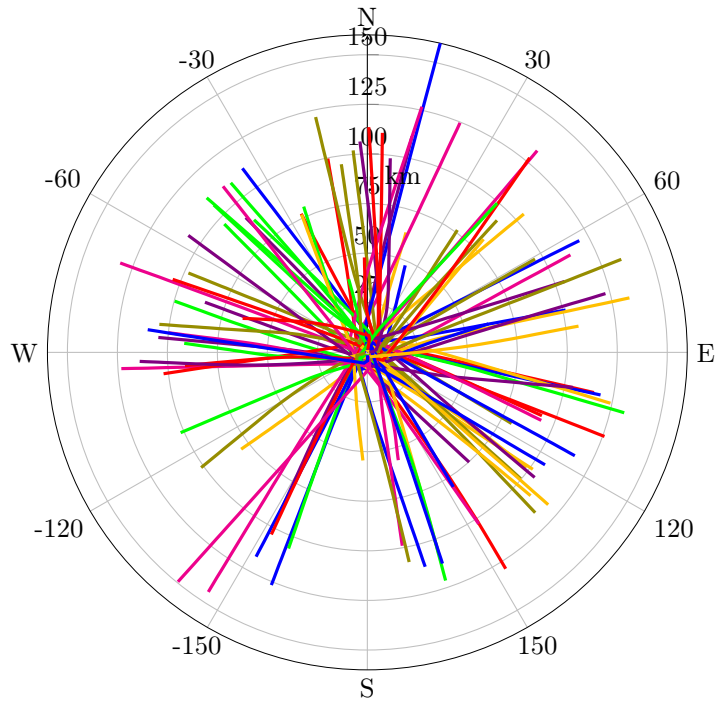

Figure \ref{fig:Sim_NSE_Results_xhor_radar} shows the horizontal tracks obtained with the aid of GNSS when executing the \nm{\nEX = 100} runs of scenario \#1. Note that the variability in the flown distance is caused by both the airspeed variations shown in table \ref{tab:Sim_Scenario_Mission_variables} as well as those of the wind field described in table \ref{tab:Sim_Scenario_Weather_variables} below.


\subsubsection*{Weather and Wind} 

\begin{figure}[h]
\centering
\begin{tikzpicture}[auto,>=latex',scale=1.8]
	\coordinate (origin)     at (+0.0,+0.0);
	\coordinate (tGNSS)      at (+0.5,+0.0);
	\coordinate (tINIDeltaT) at (+2.0,+0.0);
	\coordinate (tENDDeltaT) at (+5.9,+0.0);
	\coordinate (tINIDeltap) at (+2.6,+0.0);
	\coordinate (tENDDeltap) at (+4.7,+0.0);
	\coordinate (tINIWIND)   at (+3.3,+0.0);
	\coordinate (tENDWIND)   at (+6.8,+0.0);
	\coordinate (tEND)       at (+8.0,+0.0);
	
	\coordinate (DeltaTINI)  at (+0.0,+0.2);
	\coordinate (DeltaTEND)  at (+0.0,+2.7);
	\coordinate (DeltapINI)  at (+0.0,+2.4);
	\coordinate (DeltapEND)  at (+0.0,+0.9);
	\coordinate (vwindINI)   at (+0.0,+0.6);
	\coordinate (vwindEND)   at (+0.0,+2.2);
	\coordinate (chiWINDINI) at (+0.0,+1.6);
	\coordinate (chiWINDEND) at (+0.0,+1.3);
	\coordinate (yEND)       at (+0.0,+2.9);
	
	\coordinate (corner)       at ($(tEND)       + (yEND)$);
	\coordinate (XXDeltaTINI)  at ($(tINIDeltaT) + (DeltaTINI)$);
	\coordinate (XXDeltaTEND)  at ($(tENDDeltaT) + (DeltaTEND)$);
	\coordinate (XXDeltaT111)  at ($(tEND)       + (DeltaTEND)$);
	\coordinate (YYDeltapINI)  at ($(tINIDeltap) + (DeltapINI)$);
	\coordinate (YYDeltapEND)  at ($(tENDDeltap) + (DeltapEND)$);
	\coordinate (YYDeltap111)  at ($(tEND)       + (DeltapEND)$);
	\coordinate (ZZvwindINI)   at ($(tINIWIND)   + (vwindINI)$);
	\coordinate (ZZvwindEND)   at ($(tENDWIND)   + (vwindEND)$);
	\coordinate (ZZvwind111)   at ($(tEND)       + (vwindEND)$);
	\coordinate (TTchiWINDINI) at ($(tINIWIND)   + (chiWINDINI)$);
	\coordinate (TTchiWINDEND) at ($(tENDWIND)   + (chiWINDEND)$);
	\coordinate (TTchiWIND111) at ($(tEND)       + (chiWINDEND)$);
				
	\draw [name] [->] (origin) -- (tEND);
	\draw [name] [->] (origin) -- (yEND);
	\draw [name] [-]  (tEND) -- (corner);
	\draw [name] [-]  (yEND) -- (corner);
	\path node at($(tGNSS)     +(+0.0,-0.2)$) {\nm{\tGNSS}};
	\path node at($(tINIDeltaT)+(+0.0,-0.2)$) {\nm{\tINIDeltaT}};
	\path node at($(tENDDeltaT)+(+0.0,-0.2)$) {\nm{\tENDDeltaT}};
	\path node at($(tINIDeltap)+(+0.0,-0.2)$) {\nm{\tINIDeltap}};
	\path node at($(tENDDeltap)+(+0.0,-0.2)$) {\nm{\tENDDeltap}};
	\path node at($(tINIWIND)  +(+0.0,-0.2)$) {\nm{\tINIWIND}};
	\path node at($(tENDWIND)  +(+0.0,-0.2)$) {\nm{\tENDWIND}};	
	\path node at($(tEND)      +(+0.0,-0.2)$) {\nm{\tEND}};
	\path node at($(DeltaTINI) +(-0.3,-0.0)$) {\nm{\DeltaTINI}};
	\path node at($(DeltaTEND) +(-0.35,-0.0)$) {\nm{\DeltaTEND}};
	\path node at($(DeltapINI) +(-0.3,-0.0)$) {\nm{\DeltapINI}};
	\path node at($(DeltapEND) +(-0.35,-0.0)$) {\nm{\DeltapEND}};
	\path node at($(vwindINI)  +(-0.4,-0.0)$) {\nm{\vwindINI}};
	\path node at($(vwindEND)  +(-0.45,-0.0)$) {\nm{\vwindEND}};
	\path node at($(chiWINDINI)+(-0.45,-0.0)$) {\nm{\chiWINDINI}};
	\path node at($(chiWINDEND)+(-0.5,-0.0)$) {\nm{\chiWINDEND}};
	
	\draw [black,  ultra thin, dashed] [-] (tGNSS)      -- ($(tGNSS) + (yEND)$);
	\draw [black,  ultra thin, dashed] [-] (tINIDeltaT) -- (XXDeltaTINI);
	\draw [black,  ultra thin, dashed] [-] (tENDDeltaT) -- (XXDeltaTEND);
	\draw [black,  ultra thin, dashed] [-] (tINIDeltap) -- (YYDeltapINI);
	\draw [black,  ultra thin, dashed] [-] (tENDDeltap) -- (YYDeltapEND);
	\draw [black,  ultra thin, dashed] [-] (tINIWIND)   -- (ZZvwindINI);
	\draw [black,  ultra thin, dashed] [-] (tENDWIND)   -- (ZZvwindEND);
	\draw [black,  ultra thin, dashed] [-] (tINIWIND)   -- (TTchiWINDINI);
	\draw [black,  ultra thin, dashed] [-] (tENDWIND)   -- (TTchiWINDEND);
	\draw [black,  ultra thin, dashed] [-] (DeltaTEND)  -- (XXDeltaTEND);
	\draw [black,  ultra thin, dashed] [-] (DeltapEND)  -- (YYDeltapEND);
	\draw [black,  ultra thin, dashed] [-] (vwindEND)   -- (ZZvwindEND);
	\draw [black,  ultra thin, dashed] [-] (chiWINDEND) -- (TTchiWINDEND);
		
	\draw [blue,    name path = DeltaT, thick] [-] (DeltaTINI) -- (XXDeltaTINI) -- (XXDeltaTEND) -- (XXDeltaT111);
	\draw [red,     name path = Deltap, thick] [-] (DeltapINI) -- (YYDeltapINI) -- (YYDeltapEND) -- (YYDeltap111);
	\draw [green,   name path = vwind,  thick] [-] (vwindINI)  -- (ZZvwindINI)  -- (ZZvwindEND)  -- (ZZvwind111);
	\draw [magenta, name path = chiWIND,thick] [-] (chiWINDINI)-- (TTchiWINDINI)-- (TTchiWINDEND)-- (TTchiWIND111);
		
\end{tikzpicture}
\caption{Schematic view of scenario \#1 weather and wind field}
\label{fig:Sim_Scenario_Weather_scheme}
\end{figure}
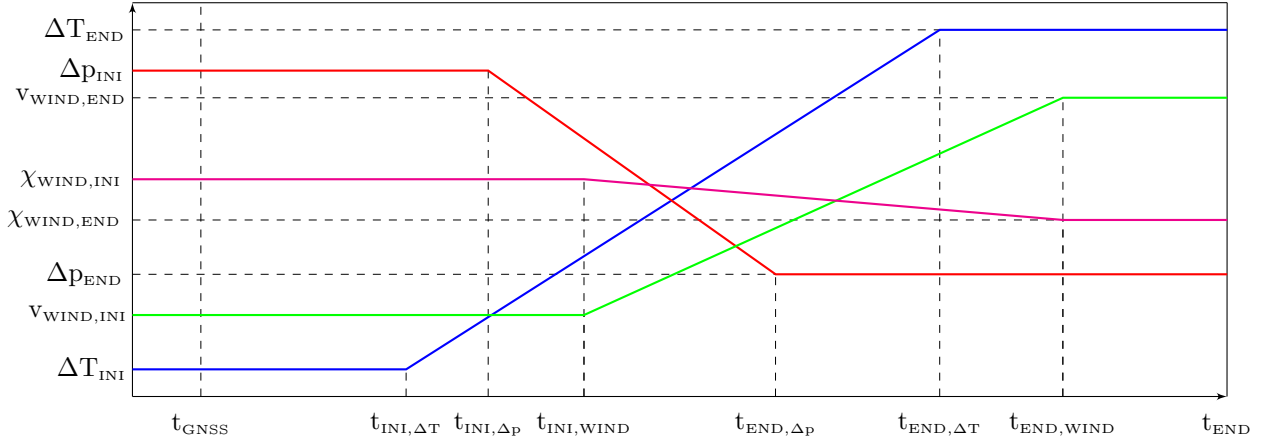

The simulation adopts the INSA model \cite{INSA} to define the atmospheric conditions encountered by the aircraft during flight, in which the temperature and pressure offsets (\nm{\DeltaT, \, \Deltap}) are provided as functions of horizontal position and time. Similarly, the low frequency wind field is also defined as two user provided functions\footnote{Both north and east wind, as well as magnitude and direction can be employed.} based on horizontal position, altitude, and time. In order to properly represent the influence that weather and wind field changes may have in the resulting GNSS-Denied trajectory, scenario \#1 is defined as containing constant weather and wind field both at the beginning (\nm{\DeltaTINI}, \nm{\DeltapINI}, \nm{\vWINDINI}) and at the end (\nm{\DeltaTEND}, \nm{\DeltapEND}, \nm{\vWINDEND}) of the trajectory, with linear variations in between, as shown in figure \ref{fig:Sim_Scenario_Weather_scheme}.

The values determining the initial and final weather and wind field, together with the times at which the linear variations start and conclude, are defined stochastically as shown in table \ref{tab:Sim_Scenario_Weather_variables}.
\begin{center}
\begin{tabular}{lcccc} 
	\hline
	\textbf{Component} & \textbf{Variable} & \textbf{Definition} & \textbf{Unit} & \textbf{Restrictions} \\
	\hline
	Temp ramp initial time         & \nm{\tINIDeltaT}    & \nm{N\lrp{400, 600^2}}                & [sec]                 & \nm{\tINIDeltaT > 50} \\
	Temp ramp final time           & \nm{\tENDDeltaT}    & \nm{N\lrp{\tINIDeltaT + 1200, 600^2}} & [sec]                 & \nm{\tENDDeltaT > \tINIDeltaT + 600} \\
	Initial temperature offset     & \nm{\DeltaTINI}     & \nm{N\lrp{0, 10^2}}                   & \nm{\lrsb{^{\circ}K}} & \\
	Final temperature offset       & \nm{\DeltaTEND}     & \nm{N\lrp{\DeltaTINI, 3^2}}           & \nm{\lrsb{^{\circ}K}} & \\
	Pressure ramp initial time     & \nm{\tINIDeltap}    & \nm{N\lrp{400, 600^2}}                & [sec]                 & \nm{\tINIDeltap > 50} \\
	Pressure ramp final time       & \nm{\tENDDeltap}    & \nm{N\lrp{\tINIDeltap + 1200, 600^2}} & [sec]                 & \nm{\tENDDeltap > \tINIDeltap + 600} \\
	Initial pressure offset        & \nm{\DeltapINI}     & \nm{N\lrp{0, 1500^2}}                 & [pa]                  & \\
	Final pressure offset          & \nm{\DeltapEND}     & \nm{N\lrp{\DeltapINI, 300^2}}         & [pa]                  & \\	
	Wind ramp initial time         & \nm{\tINIWIND}      & \nm{N\lrp{400, 600^2}}                & [sec]                 & \nm{\tINIWIND > 50} \\
	Wind ramp final time           & \nm{\tENDWIND}      & \nm{N\lrp{\tINIDeltaT + 1200, 600^2}} & [sec]                 & \nm{\tENDWIND > \tINIWIND + 300} \\
	Initial wind speed			   & \nm{\vwindINI}      & \nm{N\lrp{0, 7^2}}		   		     & [m/sec]               & \\	
	Final wind speed			   & \nm{\vwindEND}      & \nm{N\lrp{\vwindINI, 3^2}}			 & [m/sec]               &  \\
	Initial wind bearing           & \nm{\chiWINDINI}    & \nm{U\lrp{-179, 180}}				 & [deg]                 & \\
	Final wind bearing             & \nm{\chiWINDEND}    & \nm{N\lrp{\chiWINDINI, 15^2}}		 & [deg]                 & \\
\hline
\end{tabular}
\end{center}
\captionof{table}{Deterministic and stochastic components of scenario \#1 weather and wind field} \label{tab:Sim_Scenario_Weather_variables}

The values defining the weather changes have been selected by the author so they are significantly more intense than what a normal flight would experience, with the intention of highlighting any negative effects that weather and wind field changes may have in the accuracy of the GNSS-Denied navigation algorithms. On average they imply a temperature change of \nm{3 \, \lrsb{^{\circ}K}}\footnote{Note that this temperature change does not include altitude changes and is exclusively restricted to atmosphere warming or cooling. This is intended to represent positive changes during the morning and negative during the evenings.}, a pressure change of \nm{300 \, \lrsb{pa}}\footnote{As in the temperature case, this pressure change does not include altitude changes, and represents the change produced by the aircraft flying into a high or low pressure weather system. To provide more clarity, note that \nm{108600 \, \lrsb{pa}} is the highest sea level pressure ever recorded, \nm{103000 \, \lrsb{pa}} corresponds to a strong high pressure system, \nm{101325 \, \lrsb{pa}} is the standard sea level pressure \cite{ISA}, \nm{100000 \, \lrsb{pa}} corresponds to a typical low pressure system, \nm{95000 \, \lrsb{pa}} to a category III hurricane, and \nm{87000 \, \lrsb{pa}} is the lowest ever recorded.}, and a wind change of \nm{3 \, \lrsb{m/sec}} and \nm{15 \, \lrsb{deg}} in bearing, all over \nm{20 \, \lrsb{min}}.
\begin{figure}[h]
\centering
\begin{tikzpicture}
\begin{axis}[
colormap name = bluered,
cycle list={[of colormap]},
width=8.0cm,
xmin=-5000, xmax=5000, xtick={-5000,-3000,...,5000},
xlabel={\nm{\Deltap \lrsb{pa}}},
xmajorgrids,
ymin=-25, ymax=+26, ytick={-25,-15,...,25},
ylabel={\nm{\DeltaT \, \lrsb{^{\circ}K}}},
ymajorgrids,
axis lines=left,
axis line style={-stealth},
legend entries={\nm{\DeltaTINI, \, \DeltapINI}, \nm{\DeltaTEND, \, \DeltapEND}},
legend style={font=\footnotesize},
legend cell align=center,
legend pos={north west},
] 
\pgfplotstableread{figs/seeds_weather_offsets.txt}\mytable
\addplot+ [scatter, only marks, mark=*,       scatter src=explicit] table [header=false, x index=7,y index=2, meta index=0] {\mytable};
\addplot+ [scatter, only marks, mark=square*, scatter src=explicit] table [header=false, x index=9,y index=4, meta index=0] {\mytable};
\end{axis}	
\end{tikzpicture}
\hskip 10pt
\begin{tikzpicture}
\begin{axis}[
colormap name = bluered,
cycle list={[of colormap]},
width=8.0cm,
xmin=-800, xmax=1000, xtick={-800,-600,...,1000},
xlabel={\nm{\DeltapEND - \DeltapINI \lrsb{pa}}},
xmajorgrids,
ymin=-6, ymax=+6, ytick={-6,-4,...,6},
ylabel={\nm{\DeltaTEND - \DeltaTINI \, \lrsb{^{\circ}K}}},
ymajorgrids,
axis lines=left,
axis line style={-stealth},
] 
\pgfplotstableread{figs/seeds_weather_offsets.txt}\mytable
\addplot+ [scatter, only marks, mark=pentagon*,  scatter src=explicit] table [header=false, x index=10,y index=5, meta index=0] {\mytable};
\end{axis}		
\end{tikzpicture}%
\caption{Randomness in scenario \#1 temperature and pressure offsets}
\label{fig:Sim_Scenario_Weather_seeds_offsets}
\end{figure}
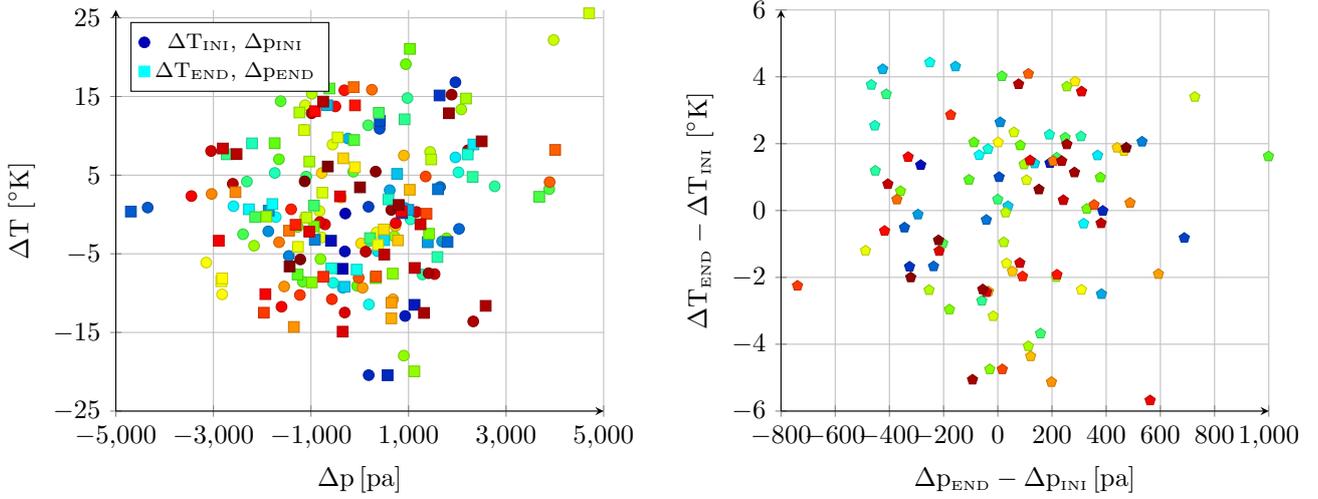

The seeds \nm{\seedWEATHER} and \nm{\seedWIND} corresponding to each trajectory seed \nm{\seed} are the only inputs required to obtain all the values listed in table \ref{tab:Sim_Scenario_Weather_variables}, hence ensuring that the results are repeatable if the same seed is employed again. Figure \ref{fig:Sim_Scenario_Weather_seeds_offsets} showcases the wide variety of weather conditions encountered during the simulation. As in previous figures, the left hand plot shows the initial and final temperature and pressure offsets (\nm{\DeltaT}, \nm{\Deltap}) for each of the \nm{\nEX} different executions, while the right hand plot shows the changes in offsets between the start and the end of the trajectory. Note that although most values can be considered normal, some executions involve weather conditions that deviate substantially from standard ones, which is the objective behind the values selected for table \ref{tab:Sim_Scenario_Weather_variables}.
\begin{figure}[h]
\centering
\begin{tikzpicture}
\begin{polaraxis}[
colormap name = bluered,
cycle list={[of colormap]},
width=7.7cm,
rotate=-90,
x dir=reverse,
axis lines*=none,
axis line style = {draw=transparent,line width=0.0001pt},
xticklabel style={anchor=-\tick-90},
xtick={0,30,60,90,120,150,180,210,240,270,300,330},
xticklabels={N,30,60,E,120,150,S,-150,-120,W,-60,-30},
ymin=0, ymax=20.0, ytick={0,5,...,15},
legend entries={\nm{\vWINDINI \, \lrsb{m/sec,deg}}, \nm{\vWINDEND \, \lrsb{m/sec,deg}}},
legend style={font=\footnotesize},
legend cell align=right,
] 
\pgfplotstableread{figs/seeds_weather_wind.txt}\mytable
\addplot+ [scatter, only marks, mark=*,       scatter src=explicit] table [header=false, x index=5,y index=3, meta index=0] {\mytable};
\addplot+ [scatter, only marks, mark=square*, scatter src=explicit] table [header=false, x index=6,y index=4, meta index=0] {\mytable};
\end{polaraxis}	
\end{tikzpicture}
\hskip 1pt
\begin{tikzpicture}
\begin{polaraxis}[
colormap name = bluered,
cycle list={[of colormap]},
width=7.7cm,
rotate=-90,
x dir=reverse,
axis lines*=none,
axis line style = {draw=transparent,line width=0.0001pt},
xticklabel style={anchor=-\tick-90},
xtick={0,30,60,90,120,150,180,210,240,270,300,330},
xticklabels={0,30,60,90,120,150,180,-150,-120,-90,-60,-30},
ymin=0, ymax=8.0, ytick={0,2,...,6},
axis line style={-stealth},
legend entries={\nm{\lvert\vWINDEND - \vWINDINI\rvert}},
legend style={font=\footnotesize},
legend cell align=right,
] 
\pgfplotstableread{figs/seeds_weather_wind.txt}\mytable
\addplot+ [scatter, only marks, mark=pentagon*,  scatter src=explicit] table [header=false, x index=8,y index=7, meta index=0] {\mytable};
\end{polaraxis}	
\end{tikzpicture}
\caption{Randomness in scenario \#1 wind speeds}
\label{fig:Sim_Scenario_Weather_seeds_wind}
\end{figure}
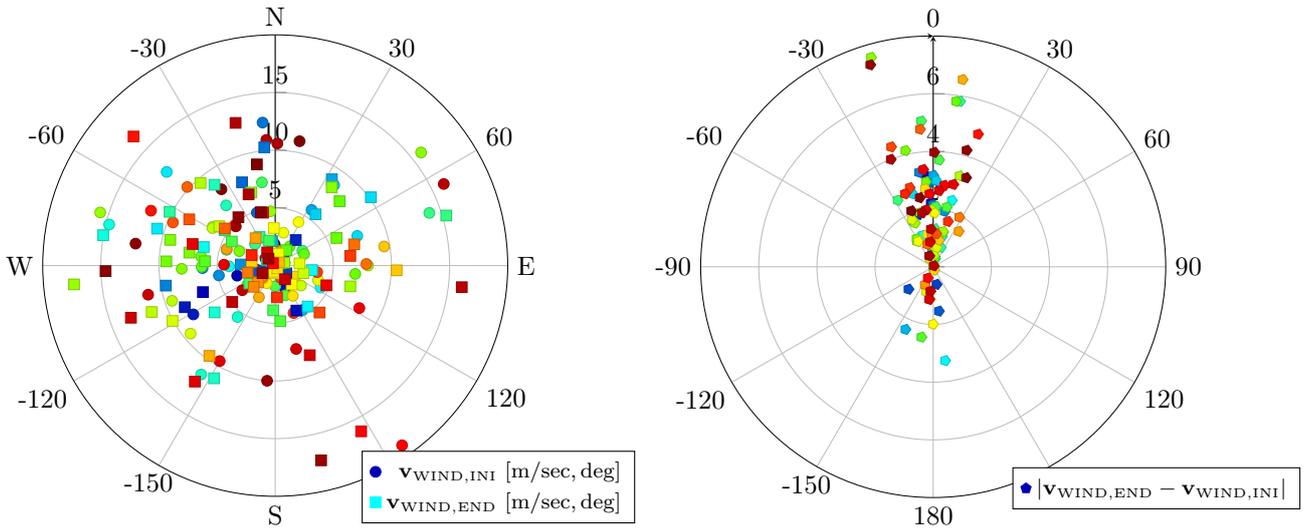

Figure \ref{fig:Sim_Scenario_Weather_seeds_wind} shows the wind field for each of the different \nm{\nEX} executions. As in previous figures, the left hand plot represents the initial and final wind speed, while the right hand plots shows the wind change experienced by the aircraft during its flight. In contrast with figure \ref{fig:Sim_Scenario_Mission_seeds_bearing}, these plots show both the wind field intensity and bearing. Note the relatively high wind speeds and wind changes experienced by some executions when compared with the aircraft airspeed values discussed in table \ref{tab:Sim_Scenario_Mission_variables}.

The stochastic nature of the scenario \#1 weather and wind fields does not only reside in their initial and final values represented in figures \ref{fig:Sim_Scenario_Weather_seeds_offsets} and \ref{fig:Sim_Scenario_Weather_seeds_wind}, but in the velocity at which the weather and wind field transition from the initial values to their final ones. Although not plotted, the velocity of these changes is specified in table \ref{tab:Sim_Scenario_Weather_variables}.


\subsubsection*{Terrain Type}

The type of terrain overflown by the aircraft has a significant influence in the performance of the visual navigation algorithms. In particular, the terrain texture (or lack of) and its elevation relief are the two most important characteristics in this regard. For this reason, each of the scenario \#1 \nm{\nEX} Monte Carlo simulation runs is repeated four times with the only change of the initial geodetic coordinates, so the resulting flights occur in four different zones or types of terrain. The zones described below are intended to represent a wide array of terrain types\footnote{Alpine terrain with elevated vertical relief is not included as piston engines aircraft can not fly high enough for the previously described \texttt{Earth Viewer} tool to generate realistic images in this type of terrain.}. Note that changing the area of the world were the flight takes place also modifies the gravitational and magnetic fields, but as all considered zones are far from the Earth magnetic poles, this does not imply any significant change in the performance of the inertial algorithms. Images representative of each zone as viewed by the onboard camera are also included.
\begin{itemize}
\item The ``\textbf{desert}'' (DS) zone is located in the Sonoran desert of southern Arizona (USA) and northern Mexico, with an initial position given by \nm{\lambda_{DS} = 248.001185 \lrsb{deg}}, \nm{\varphi_{DS} = 32.157903 \lrsb{deg}}, and \nm{h_{{\sss{GROUND}},DS} = 661.0 \lrsb{m}}. It is characterized by a combination of bajadas (broad slopes of debris) and isolated very steep mountain ranges. There is virtually no human infrastructure or flat terrain, as the bajadas have sustained slopes of up to 7 [deg]. The altitude of the bajadas ranges from 300 to 800 [m] above \texttt{MSL}, and the mountains reach up to 800 [m] above the surrounding terrain. Texture is abundant because of the cacti and the vegetation along the dry creeks.
\begin{figure}[h]
\centering
\includegraphics[width=7.5cm]{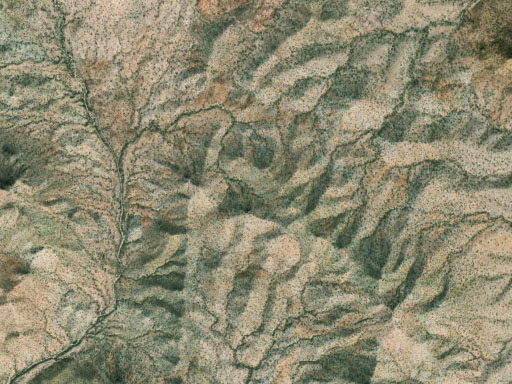}
\hskip 10pt
\includegraphics[width=7.5cm]{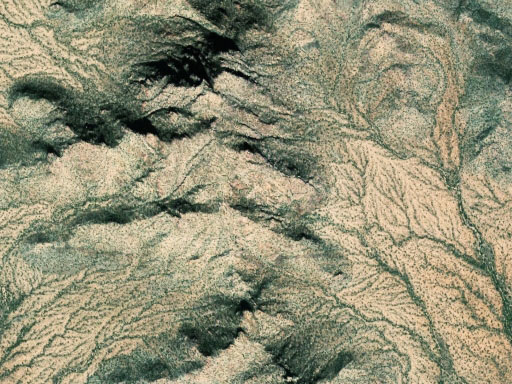}
\caption{Typical ``desert'' (DS) terrain view}
\label{fig:Sim_Scenario_Terrain_desert}
\end{figure}

\item The ``\textbf{farm}'' (FM) zone is located in the fertile farmland of southeastern Illinois and southwestern Indiana (USA), with an initial position given by \nm{\lambda_{FM} = 272.122371 \lrsb{deg}}, \nm{\varphi_{FM} = 38.865625 \lrsb{deg}}, and \nm{h_{{\sss{GROUND}},FM} = 144.0 \lrsb{m}}. A significant percentage of the terrain is made of regular plots of farmland, but there also exists some woodland, farm houses, rivers, lots of little towns, and roads. It is mostly flat with an altitude above \texttt{MSL} between 100 and 200 [m], and altitude changes are mostly restricted to the few forested areas. Texture is nonexistent in the farmlands, where extracting features is often impossible.
\begin{figure}[h]
\centering
\includegraphics[width=7.5cm]{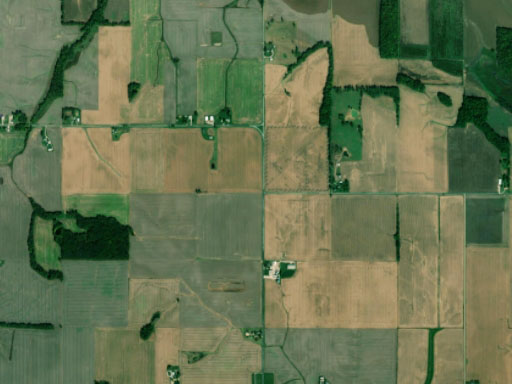}
\hskip 10pt
\includegraphics[width=7.5cm]{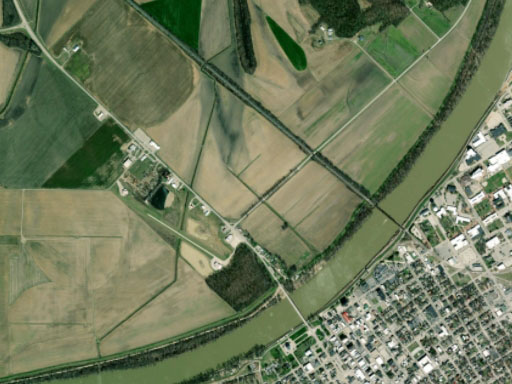}
\caption{Typical ``farm'' (FM) terrain view}
\label{fig:Sim_Scenario_Terrain_farm}
\end{figure}

\item The ``\textbf{forest}'' (FR) zone is located in the deciduous forestlands of Vermont and New Hamspshire (USA), with an initial position given by \nm{\lambda_{FR} = 287.490805 \lrsb{deg}}, \nm{\varphi_{FR} = 43.354486 \lrsb{deg}}, and \nm{h_{{\sss{GROUND}},FR} = 200.0 \lrsb{m}}. The terrain is made up of forests and woodland, with some clearcuts, small towns, and roads. There are virtually no flat areas, as the land is made up by hills and small to medium size mountains that are never very steep. The valleys range from 100 to 300 [m] above \texttt{MSL}, while the tops of the mountains reach 500 to 900 [m]. Features are plentiful in the woodlands.
\begin{figure}[h]
\centering
\includegraphics[width=7.5cm]{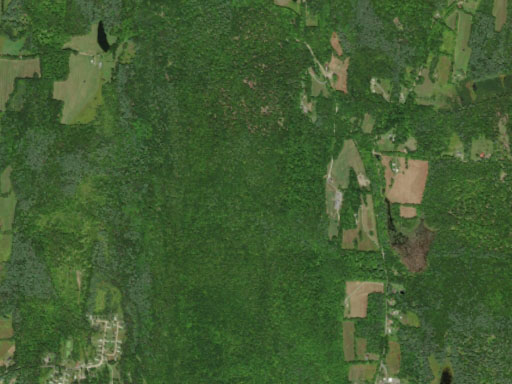}
\hskip 10pt
\includegraphics[width=7.5cm]{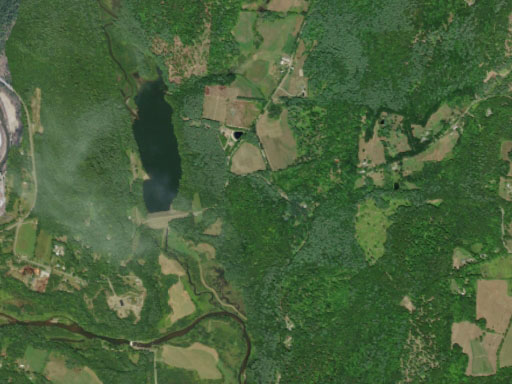}
\caption{Typical ``forest'' (FR) terrain view}
\label{fig:Sim_Scenario_Terrain_forest}
\end{figure}

\item The ``\textbf{mix}'' (MX) zone is located in northern Mississippi and extreme southwestern Tennessee (USA), with an initial position given by \nm{\lambda_{MX} = 270.984538 \lrsb{deg}}, \nm{\varphi_{MX} = 34.720636 \lrsb{deg}}, and \nm{h_{{\sss{GROUND}},MX} = 133.0 \lrsb{m}}. Approximately half of the land consists of woodland in the hills, and the other half is made up by farmland in the valleys, with a few small towns and roads. Altitude changes are always presents and the terrain is never flat, but they are smaller than in the DS and FR zones, with the altitude oscillating between 100 and 200 [m] above \texttt{MSL}.
\begin{figure}[h]
\centering
\includegraphics[width=7.5cm]{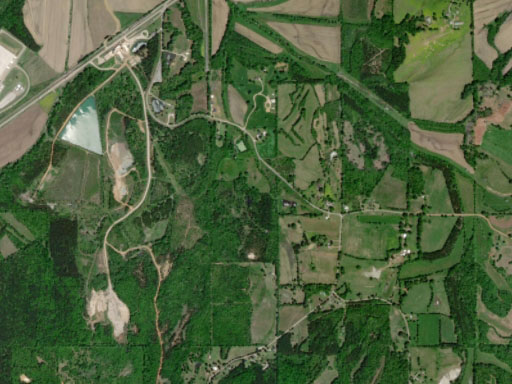}
\hskip 10pt
\includegraphics[width=7.5cm]{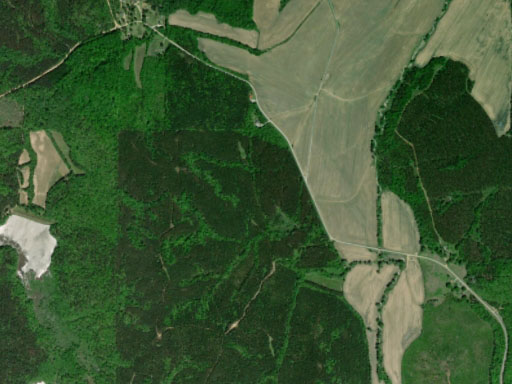}
\caption{Typical ``mix'' (MX) terrain view}
\label{fig:Sim_Scenario_Terrain_mix}
\end{figure}
\end{itemize}


\section{Scenario \#2}\label{sec:Scenario2}

Scenario \#1 is intended to represent the challenges involved in reaching a predefined recovery location without the usage of GNSS signals and while experiencing dynamic weather and wind conditions. Although it is quite generic, there are two reasons why it can not constitute the only scenario employed to evaluate the behavior of different GNSS-Denied navigation algorithms. On one side, although the scenario \#1 mission includes three maneuvers (a change of bearing, a change of airspeed, and a change of pressure altitude), for the most part it consists of a long straight level flight in which the aircraft systems have lots of time to recover from estimation errors induced by the maneuvers, which are executed quite apart from each other. Scenario \#2 includes several maneuvers and avoids long straight level segments to ensure that the GNSS-Denied navigation algorithms are valid for different types of missions.

Additionally, the GNSS-Denied position errors incurred by some inertial navigation systems may exhibit a significant dependency on the weather and wind changes that occur since the aircraft can no longer employ the GNSS signals for navigation. The prevalence of these errors coupled with the significant scenario \#1 weather and wind changes prevents the detection of smaller error sources when analyzing the Monte Carlo simulation results. 

These two reasons indicate the need for a scenario \nm{\#2}, which represents the challenges involved in continuing with the original mission upon the loss of the GNSS signals, executing a series of continuous maneuvers over a relatively short period of time with no weather or wind variations.


\subsubsection*{Mission}

The mission, guidance targets, or reference trajectory of scenario \#2 consists in the seventeen flight segments defined in (\ref{eq:Sim_Scenario_Alter_Mission_RT}) through (\ref{eq:Sim_Scenario_Alter_Mission_target17}) below. It is intended to showcase the ability of an autonomous aircraft to continue executing its mission while in GNSS-Denied conditions, specially in a congested, urban, or mountainous environment in which continuous maneuvers are required to avoid other aircraft, infrastructure, or terrain.
\begin{itemize}

\item The whole trajectory is executed at constant airspeed \nm{\vtasINI}, constant pressure altitude \nm{\HpINI}, and no sideslip \nm{\beta = 0}. 

\item The first segment coincides with that of scenario \#1, as the aircraft flies at constant bearing \nm{\chiINI}, first with the assistance of GNSS signals until \nm{\tGNSS = 100 \lrsb{sec}} and then in GNSS-Denied conditions until \nm{t_{{\sss TURN},1}}.

\item The rest of the mission consists in a series of eight turning maneuvers connecting straight segments flown at constant bearings \nm{\chi_i}. Turns are executed with body bank angles of \nm{\xiTURNi = \pm 10 \, \lrsb{deg}}, with the sign given by the shortest turn necessary to reach the next bearing \nm{\chi_i}, upon which the turn concludes, and the duration of the straight segments is given by \nm{\DeltatTURNi}. The short time between consecutive maneuvers is intended to avoid providing the inertial navigation system with sufficient time to stabilize itself in case the maneuvers have a destabilizing effect on the estimation errors.

\item The scenario concludes at \nm{\tEND = 500 \, \lrsb{sec}}, which for some seeds may not be sufficient to conclude the eight turns.
\end{itemize}
\begin{eqnarray}
\nm{\xREF}                 & = & \nm{\lrb{\deltaTARGETone, \, \deltaTARGETtwo, \dots, \deltaTARGETseventeen}} \label{eq:Sim_Scenario_Alter_Mission_RT} \\
\nm{\deltaTARGETone}       & = & \nm{\lrsb{\vtasINI, \, \HpINI, \,  \chiINI, \, \beta = 0, \, t_{{\sss TURN},1}}^T}\label{eq:Sim_Scenario_Alter_Mission_target1} \\
\nm{\deltaTARGETtwo}       & = & \nm{\lrsb{\vtasINI, \, \HpINI, \,  \xi_1,   \, \beta = 0, \, \chi_1}^T}\label{eq:Sim_Scenario_Alter_Mission_target2} \\
\nm{\deltaTARGETthr}       & = & \nm{\lrsb{\vtasINI, \, \HpINI, \,  \chi_1,  \, \beta = 0, \, \Delta t_{{\sss TURN},2}}^T}\label{eq:Sim_Scenario_Alter_Mission_target3} \\
\nm{\cdots}                &   & \nm{\cdots}\nonumber \\
\nm{\deltaTARGETfourteen}  & = & \nm{\lrsb{\vtasINI, \, \HpINI, \,  \xi_7,   \, \beta = 0, \, \chi_7}^T}\label{eq:Sim_Scenario_Alter_Mission_target14} \\
\nm{\deltaTARGETfifteen}   & = & \nm{\lrsb{\vtasINI, \, \HpINI, \,  \chi_7,  \, \beta = 0, \, \Delta t_{{\sss TURN},8}}^T}\label{eq:Sim_Scenario_Alter_Mission_target15} \\
\nm{\deltaTARGETsixteen}   & = & \nm{\lrsb{\vtasINI, \, \HpINI, \,  \xi_8,   \, \beta = 0, \, \chi_8}^T}\label{eq:Sim_Scenario_Alter_Mission_target16} \\
\nm{\deltaTARGETseventeen} & = & \nm{\lrsb{\vtasINI, \, \HpINI, \,  \chi_8,  \, \beta = 0, \, \tEND}^T}\label{eq:Sim_Scenario_Alter_Mission_target17} 
\end{eqnarray}

Table \ref{tab:Sim_Scenario_Alter_Mission_variables} shows the definition of the deterministic and stochastic variables on which the scenario \#2 mission relies. Note that it has been developed with the objective of including as much variability as possible, as the only deterministic variables are those that do not result in loss of generality. The initial bearing, altitude, and airspeed coincide with those of scenario \#1 shown in the left hand side of figures \ref{fig:Sim_Scenario_Mission_seeds_bearing} and \ref{fig:Sim_Scenario_Mission_seeds_alt_speed}, represented by circles.
\begin{center}
\begin{tabular}{lccccc}
	\hline
	\textbf{Component} & \textbf{Variable} & \textbf{Index} & \textbf{Definition} & \textbf{Unit} & \textbf{Restrictions} \\
	\hline
	Airspeed              & \nm{\vtasINI}          &                      & \nm{N\lrp{29, 1.5^2}}     & [m/sec] & \nm{24 < \vtasINI < 34} \\
	Pressure altitude	  & \nm{\HpINI}            &                      & \nm{N\lrp{2700, 200^2}}   & [m]     & \\
	Initial bearing       & \nm{\chiINI = \chi_0}  &                      & \nm{U\lrp{-179, 180}}     & [deg]   & \\
	Sideslip angle        & \nm{\beta}             &                      & \nm{0}                    & [deg]   & \\
	GNSS-Denied time	  & \nm{\tGNSS}		       &                      & \nm{100}		   		     & [sec]   & \\
	Turn bank angles	  & \nm{\xiTURNi}	       & \nm{i = 1, \dots, 8} & \nm{\pm 10}			     & [deg]   & \\
	Intermediate bearings & \nm{\chi_i}            & \nm{i = 1, \dots, 7} & \nm{U\lrp{-179, 180}}     & [deg]   & \nm{\lvert \chi_i - \chi_{i-1} \rvert > 10} \\
    Final bearing         & \nm{\chiEND = \chi_8}  &                      & \nm{U\lrp{-179, 180}}     & [deg]   & \nm{\lvert \chi_8 - \chi_7 \rvert > 10} \\
	Initial turn time	  & \nm{t_{{\sss TURN},1}} &                      & \nm{\tGNSS + N(30, 50^2)} & [sec]   & \nm{\tTURN > \lrp{\tGNSS + 15}} \\	
	Turn time intervals   & \nm{\DeltatTURNi}      & \nm{i = 2, \dots, 8} & \nm{U\lrp{10, 50}}        & [sec]   & \\
	Final time            & \nm{\tEND}             &                      & \nm{500}					 & [sec]   & \\
	\hline
\end{tabular}
\end{center}
\captionof{table}{Deterministic and stochastic components of the scenario \#2 mission} \label{tab:Sim_Scenario_Alter_Mission_variables}
\begin{figure}[h]
\centering
\begin{tikzpicture}
\begin{polaraxis}[
cycle list={{red,no markers,very thick},
           {blue,no markers,very thick},
		   {orange!50!yellow,no markers,very thick},
		   {violet,no markers,very thick},
		   {green,no markers,very thick},
		   {magenta,no markers,very thick},
		   {olive,no markers,very thick}},
width=10.0cm,
rotate=-90,
x dir=reverse,
axis lines*=none,
axis line style = {draw=transparent,line width=0.0001pt},
xticklabel style={anchor=-\tick-90},
xtick={0,30,60,90,120,150,180,210,240,270,300,330},
xticklabels={N,30,60,E,120,150,S,-150,-120,W,-60,-30},
ymin=0,
ymax=15,
ytick={0,2,...,14},
ylabel=km,
axis line style={-stealth},
] 
\pgfplotstableread{figs/error_filter_pos_alter_hor_m.txt}\mytable
\addplot table [header=false, x index= 1,y index= 0] {\mytable};
\addplot table [header=false, x index= 5,y index= 4] {\mytable};
\addplot table [header=false, x index= 9,y index= 8] {\mytable};
\addplot table [header=false, x index=13,y index=12] {\mytable};
\addplot table [header=false, x index=17,y index=16] {\mytable}; 

\addplot table [header=false, x index=21,y index=20] {\mytable};
\addplot table [header=false, x index=25,y index=24] {\mytable};
\addplot table [header=false, x index=29,y index=28] {\mytable};
\addplot table [header=false, x index=33,y index=32] {\mytable};
\addplot table [header=false, x index=37,y index=36] {\mytable}; 

\addplot table [header=false, x index=41,y index=40] {\mytable};
\addplot table [header=false, x index=45,y index=44] {\mytable};
\addplot table [header=false, x index=49,y index=48] {\mytable};
\addplot table [header=false, x index=53,y index=52] {\mytable};
\addplot table [header=false, x index=57,y index=56] {\mytable};

\addplot table [header=false, x index=61,y index=60] {\mytable};
\addplot table [header=false, x index=65,y index=64] {\mytable};
\addplot table [header=false, x index=69,y index=68] {\mytable};
\addplot table [header=false, x index=73,y index=72] {\mytable};
\addplot table [header=false, x index=77,y index=76] {\mytable}; 

\addplot table [header=false, x index=81,y index=80] {\mytable};	
\addplot table [header=false, x index=85,y index=84] {\mytable};	
\addplot table [header=false, x index=89,y index=88] {\mytable};
\addplot table [header=false, x index=93,y index=92] {\mytable};	
\addplot table [header=false, x index=97,y index=96] {\mytable}; 

\addplot table [header=false, x index=101,y index=100] {\mytable};	
\addplot table [header=false, x index=105,y index=104] {\mytable};	
\addplot table [header=false, x index=109,y index=108] {\mytable};	
\addplot table [header=false, x index=113,y index=112] {\mytable};	
\addplot table [header=false, x index=117,y index=116] {\mytable};	 

\addplot table [header=false, x index=121,y index=120] {\mytable};	
\addplot table [header=false, x index=125,y index=124] {\mytable};	
\addplot table [header=false, x index=129,y index=128] {\mytable};	
\addplot table [header=false, x index=133,y index=132] {\mytable};
\addplot table [header=false, x index=137,y index=136] {\mytable};	 

\addplot table [header=false, x index=141,y index=140] {\mytable};	
\addplot table [header=false, x index=145,y index=144] {\mytable};	
\addplot table [header=false, x index=149,y index=148] {\mytable};	
\addplot table [header=false, x index=153,y index=152] {\mytable};	
\addplot table [header=false, x index=157,y index=156] {\mytable};	 

\addplot table [header=false, x index=161,y index=160] {\mytable};	
\addplot table [header=false, x index=165,y index=164] {\mytable};	
\addplot table [header=false, x index=169,y index=168] {\mytable};	
\addplot table [header=false, x index=173,y index=172] {\mytable};	
\addplot table [header=false, x index=177,y index=176] {\mytable};	 

\addplot table [header=false, x index=181,y index=180] {\mytable};	
\addplot table [header=false, x index=185,y index=184] {\mytable};	
\addplot table [header=false, x index=189,y index=188] {\mytable};	
\addplot table [header=false, x index=193,y index=192] {\mytable};	
\addplot table [header=false, x index=197,y index=196] {\mytable};	 

\addplot table [header=false, x index=201,y index=200] {\mytable};	
\addplot table [header=false, x index=205,y index=204] {\mytable};	
\addplot table [header=false, x index=209,y index=208] {\mytable};	
\addplot table [header=false, x index=213,y index=212] {\mytable};	
\addplot table [header=false, x index=217,y index=216] {\mytable};	

\addplot table [header=false, x index=221,y index=220] {\mytable};	
\addplot table [header=false, x index=225,y index=224] {\mytable};	
\addplot table [header=false, x index=229,y index=228] {\mytable};	
\addplot table [header=false, x index=233,y index=232] {\mytable};	
\addplot table [header=false, x index=237,y index=236] {\mytable};	 

\addplot table [header=false, x index=241,y index=240] {\mytable};
\addplot table [header=false, x index=245,y index=244] {\mytable};	
\addplot table [header=false, x index=249,y index=248] {\mytable};	
\addplot table [header=false, x index=253,y index=252] {\mytable};	
\addplot table [header=false, x index=257,y index=256] {\mytable};	 

\addplot table [header=false, x index=261,y index=260] {\mytable};	
\addplot table [header=false, x index=265,y index=264] {\mytable};
\addplot table [header=false, x index=269,y index=268] {\mytable};	
\addplot table [header=false, x index=273,y index=272] {\mytable};
\addplot table [header=false, x index=277,y index=276] {\mytable};	 

\addplot table [header=false, x index=281,y index=280] {\mytable};	
\addplot table [header=false, x index=285,y index=284] {\mytable};	
\addplot table [header=false, x index=289,y index=288] {\mytable};	
\addplot table [header=false, x index=293,y index=292] {\mytable};	
\addplot table [header=false, x index=297,y index=296] {\mytable}; 

\addplot table [header=false, x index=301,y index=300] {\mytable};
\addplot table [header=false, x index=305,y index=304] {\mytable};	
\addplot table [header=false, x index=309,y index=308] {\mytable};	
\addplot table [header=false, x index=313,y index=312] {\mytable};	
\addplot table [header=false, x index=317,y index=316] {\mytable}; 

\addplot table [header=false, x index=321,y index=320] {\mytable};
\addplot table [header=false, x index=325,y index=324] {\mytable};	
\addplot table [header=false, x index=329,y index=328] {\mytable};	
\addplot table [header=false, x index=333,y index=332] {\mytable};	
\addplot table [header=false, x index=337,y index=336] {\mytable};	 

\addplot table [header=false, x index=341,y index=340] {\mytable};	
\addplot table [header=false, x index=345,y index=344] {\mytable};	
\addplot table [header=false, x index=349,y index=348] {\mytable};	
\addplot table [header=false, x index=353,y index=352] {\mytable};	
\addplot table [header=false, x index=357,y index=356] {\mytable};	

\addplot table [header=false, x index=361,y index=360] {\mytable};	
\addplot table [header=false, x index=365,y index=364] {\mytable};	
\addplot table [header=false, x index=369,y index=368] {\mytable};
\addplot table [header=false, x index=373,y index=372] {\mytable};	
\addplot table [header=false, x index=377,y index=376] {\mytable};	 

\addplot table [header=false, x index=381,y index=380] {\mytable};	
\addplot table [header=false, x index=385,y index=384] {\mytable};	
\addplot table [header=false, x index=389,y index=388] {\mytable};	
\addplot table [header=false, x index=393,y index=392] {\mytable};	
\addplot table [header=false, x index=397,y index=396] {\mytable};	 

\end{polaraxis}	
\end{tikzpicture}
\caption{GNSS-Based horizontal tracks for scenario \#2 seeds \nm{\Upsilon_{1}} through \nm{\Upsilon_{100}}}
\label{fig:Sim_NSE_Results_xhor_radar_alter}
\end{figure}
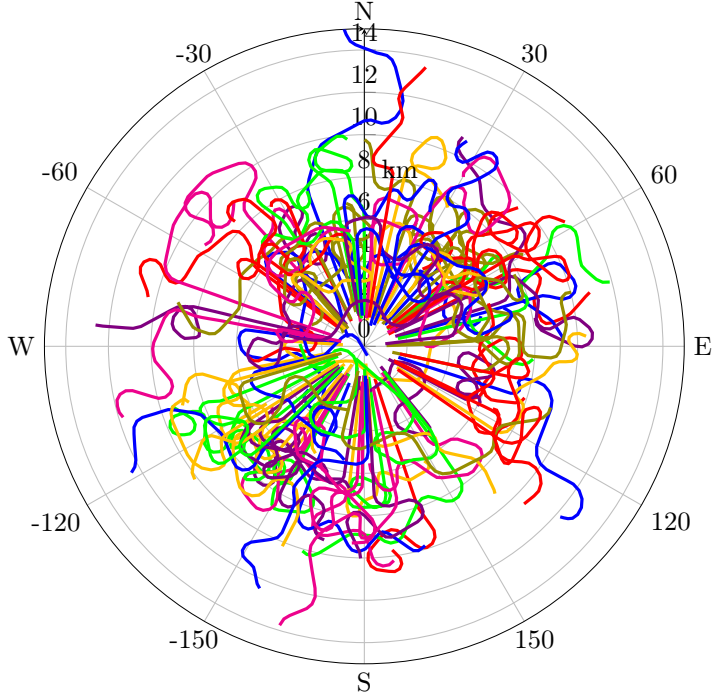

Figure \ref{fig:Sim_NSE_Results_xhor_radar_alter} shows the horizontal tracks obtained with the aid of GNSS when executing the \nm{\nEX = 100} runs of scenario \#2. 


\subsubsection*{Weather and Wind}

The influence of the weather and wind changes since the time the GNSS signals are lost is properly captured by scenario \#1. In scenario \#2, the weather and wind remain constant throughout the trajectory, which enables the identification of other navigation error sources that are obscured in scenario \#1 by the prevalence of the weather and wind changes. In contrast with scenario \#1, which was designed to represent with as much realism as possible the challenges encountered by an autonomous aircraft trying to reach a predefined recovery location, scenario \#2 is unrealistic as it completely eliminates the most important error sources for the inertial estimation of the aircraft position. A minimum amount of weather and wind changes is always present in real flights, so employing scenario \#2 will always results in predictions that are optimistic.
\begin{center}
\begin{tabular}{lccc}
	\hline
	\textbf{Component} & \textbf{Variable} & \textbf{Definition} & \textbf{Unit} \\
	\hline
	Temperature offset     & \nm{\DeltaTINI}     & \nm{N\lrp{0, 10^2}}		& \nm{\lrsb{^{\circ}K}} \\
	Pressure offset        & \nm{\DeltapINI}     & \nm{N\lrp{0, 1500^2}}	& [pa]                  \\
	Wind speed			   & \nm{\vwindINI}      & \nm{N\lrp{0, 7^2}}		& [m/sec]               \\	
	Wind bearing           & \nm{\chiWINDINI}    & \nm{U\lrp{-179, 180}}	& [deg]                 \\
\hline
\end{tabular}
\end{center}
\captionof{table}{Deterministic and stochastic components of scenario \#2 weather and wind field} \label{tab:Sim_Scenario_Alter_Weather_variables}

Scenario \#2 hence is defined as containing constant weather and wind field (\nm{\DeltaTINI}, \nm{\DeltapINI}, \nm{\vWINDINI}). These values are determined stochastically as shown in table \ref{tab:Sim_Scenario_Alter_Weather_variables}, and coincide with the initial conditions of scenario \#1 (table \ref{tab:Sim_Scenario_Weather_variables}). These are shown in the left hand side of figures \ref{fig:Sim_Scenario_Weather_seeds_offsets} and \ref{fig:Sim_Scenario_Weather_seeds_wind}, represented by circles.


\subsubsection*{Terrain Type}

The short duration and continuous maneuvering of scenario \#2 enables the use of two additional zones or terrain types in its Monte Carlo simulations. These two zones are not employed in scenario \#1 because the author could not locate wide enough areas with a prevalence of this type of terrain (note that scenario \#1 trajectories can conclude up to 125 [km] in any direction from its initial coordinates, but only 12 [km] for scenario \#2, as shown in figures \ref{fig:Sim_NSE_Results_xhor_radar} and \ref{fig:Sim_NSE_Results_xhor_radar_alter}).
\begin{itemize}
\item The ``\textbf{prairie}'' (PR) zone is located in the Everglades floodlands of southern Florida (USA), with an initial position given by \nm{\lambda_{PR} = 279.088834 \lrsb{deg}}, \nm{\varphi_{PR} = 25.855172 \lrsb{deg}}, and \nm{h_{{\sss{GROUND}},PR} = 10.0 \lrsb{m}}. It consists of flat grasslands, swamps, and tree islands located a few meters above \texttt{MSL}, with the only human infrastructure being a few dirt roads and landing strips, but no settlements. Features may be difficult to obtain in some areas due to the lack of texture.
\begin{figure}[h]
\centering
\includegraphics[width=7.5cm]{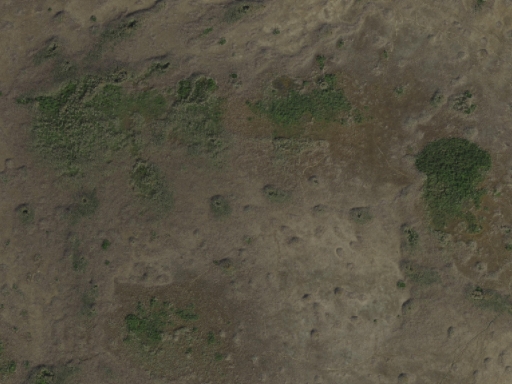}
\hskip 10pt
\includegraphics[width=7.5cm]{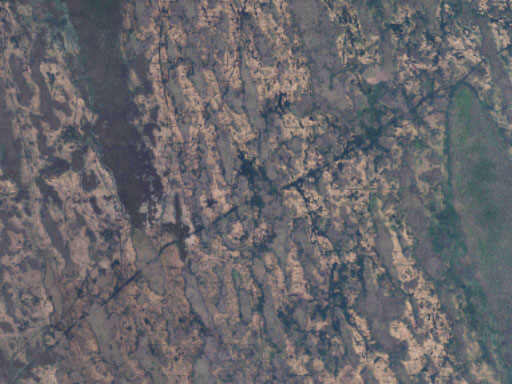}
\caption{Typical ``prairie'' (PR) terrain view}
\label{fig:Sim_Scenario_Terrain_prairie}
\end{figure}

\item The ``\textbf{urban}'' (UR) zone is located in the Los Angeles metropolitan area, with an initial position given by \nm{\lambda_{UR} = 241.799731 \lrsb{deg}}, \nm{\varphi_{UR} = 33.924426 \lrsb{deg}}, and \nm{h_{{\sss{GROUND}},UR} = 26.0 \lrsb{m}}. It is composed by a combination of single family houses and commercial buildings separated by freeways and streets. There is some vegetation but no natural landscapes, and the terrain is flat and close to \texttt{MSL}.
\begin{figure}[h]
\centering
\includegraphics[width=7.5cm]{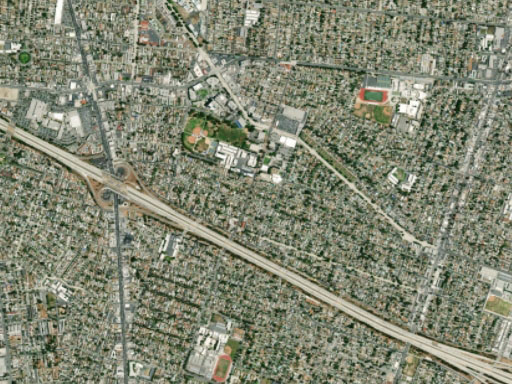}
\hskip 10pt
\includegraphics[width=7.5cm]{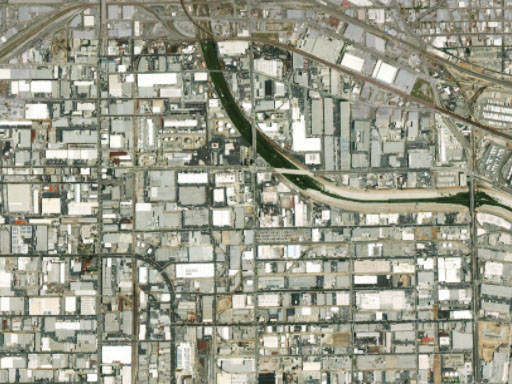}
\caption{Typical ``urban'' (UR) terrain view}
\label{fig:Sim_Scenario_Terrain_urban}
\end{figure}

\end{itemize}

Although scenario \#2 also makes use of the four terrain types listed for scenario \#1 (DS, FM, FR, and MX), it is worth noting that the variability of the terrain is significantly higher for scenario \#1 because of the bigger land extension covered. The altitude relief, abundance or scarcity of features, land use diversity, and presence of rivers and mountains is hence more varied when executing a given run of scenario \#1 on a certain type of terrain, than when executing the same run for scenario \#2. From the point of view of the influence of the terrain on the visual navigation algorithms, scenario \#1 should theoretically be more challenging than \#2.


\section{Metrics} \label{sec:Metrics}

This section defines the metrics employed to evaluate the performance of a given navigation algorithm in GNSS-Denied conditions based on the results obtained in Monte Carlo simulations applied to the two scenarios defined above. To be useful, the metrics employed must be as simple as possible so as to facilitate the comparison process between two different algorithms, but also be representative enough of the performances that the navigation system should exhibit. 

Let's imagine an error function \nm{f_X \lrp{\xvecest, \, \xvec, \, t} = f_X \lrp{\xEST, \, \xTRUTH, \, t}} that evaluates the difference between a certain variable of the observed trajectory \nm{X_{\sss EST}} and the same variable in the actual trajectory \nm{X_{\sss TRUTH}}, both at the same time t:
\neweq{f_X \lrp{\xvecest, \, \xvec, \, t} = f_X \lrp{t} = X_{\sss EST}\lrp{t} - X_{\sss TRUTH}\lrp{t}} {eq:Sim_Metrics_function_f}

This error function is evaluated simultaneously with the navigation system outputs (\nm{t = t_n = n \cdot \DeltatEST}). A good navigation system is that which diminishes the navigation system error, so the output of the error function \nm{f_X} is as small as possible. Note that for each scenario \#1 execution or seed \nm{\seed}, there exist \nm{\nEND = \tEND / \DeltatEST + 1 = 380001} evaluations of the error function \nm{f_X}, which is renamed \nm{f_{Xj}} to specify that it applies to execution \nm{\seed}. In the case of scenario \#2, the number of evaluations is \nm{50001}.
\neweq{f_{Xj} \lrp{\xvecest_j, \, \xvec_j, \, t_n} = f_{Xj} \lrp{t_n} = X_{{\sss EST}j}\lrp{t_n} - X_{{\sss TRUTH}j}\lrp{t_n}} {eq:Sim_Metrics_function2_f}


\subsubsection*{Trajectory Metrics}

The error function \nm{f_{Xj}} can be considered a random or stochastic process, and hence its mean \nm{\muj{\hat{X}}} and standard deviation \nm{\sigmaj{\hat{X}}} can be readily computed. In addition, it is also interesting to obtain \nm{\maxj{\hat{X}}}, this is, the maximum absolute value\footnote{The maximum absolute value metric of a group of values returns the signed value with the highest absolute value.} of the function \nm{f_{Xj}}:
\neweq{\begin{bmatrix} \nm{\muj{\hat{X}}} \\ \nm{\sigmaj{\hat{X}}} \\ \nm{\maxj{\hat{X}}} \end{bmatrix} = \begin{bmatrix} \nm{\cfrac{\displaystyle\sum_{n=1}^{\nEND} f_{Xj} \lrp{t_n}}{\nEND}} \ & \ \nm{\sqrt{\cfrac{\displaystyle\sum_{n=1}^{\nEND} \Big(f_{Xj} \lrp{t_n} - \muj{\hat{X}}\Big)^2}{\nEND}}} \ & \ \nm{\displaystyle\max_{n=1}^{\nEND} \ \lvert f_{Xj} \lrp{t_n}\rvert} \end{bmatrix}^T} {eq:Sim_Metrics_mean_std_max_j}

The above mean, standard deviation, and absolute value maximum provide a good understanding of how good the navigation system is at continuously estimating \nm{X_{\sss EST}} when compared with its real value \nm{X_{\sss TRUTH}} for an specific trajectory execution or seed \nm{\seed}, and for that reason they are known as \emph{trajectory metrics} or \emph{seed metrics}. 


\subsubsection*{Final State Value}

In some ocassions it may be necessary to ignore the \nm{f_{Xj}} error function evolution with time to only focus on its value at the end of the scenario (\nm{t_n = \tEND}). This value is provided by the \emph{final state value}:
\neweq{\XENDj{\hat{X}} = f_{Xj}\lrp{\tEND}} {eq:Sim_metrics_final_traj}


\subsubsection*{Aggregated Metrics}

The behavior of the navigation system can only be properly evaluated when it is applied to a sufficiently big number of executions (\nm{\nEX}), and it is for this purpose that the \emph{aggregated metrics} (means, standard deviations, and absolute value maximums) are computed\footnote{Note that the aggregated mean of maximums and the aggregated standard deviation of maximums only make sense if applied to the absolute value of the maximums instead of the maximum themselves. They are defined accordingly in (\ref{eq:Sim_Metrics_mean_mean_std_max}, \ref{eq:Sim_Metrics_std_mean_std_max}).}. To obtain the aggregated metrics, the trajectory mean \nm{\muj{\hat{X}}}, standard deviation \nm{\sigmaj{\hat{X}}}, and absolute value maximum \nm{\maxj{\hat{X}}} for each seed are considered realizations of a random variable:
\begin{eqnarray}        
\nm{\begin{bmatrix} \nm{\mumu{\hat{X}}} \\ \nm{\musigma{\hat{X}}} \\ \nm{\mumax{\hat{X}}} \end{bmatrix}} & = & \nm{\begin{bmatrix} \nm{\cfrac{\displaystyle\sum_{j=1}^{\nEX} \muj{\hat{X}}}{\nEX}} & \nm{\cfrac{\displaystyle\sum_{j=1}^{\nEX} \sigmaj{\hat{X}}}{\nEX}} & \nm{\cfrac{\displaystyle\sum_{j=1}^{\nEX}{\lvert \maxj{\hat{X}} \rvert}}{\nEX}} \end{bmatrix}^T} \label{eq:Sim_Metrics_mean_mean_std_max} \\
\nm{\begin{bmatrix} \nm{\sigmamu{\hat{X}}} \\ \nm{\sigmasigma{\hat{X}}} \\ \nm{\sigmamax{\hat{X}}} \end{bmatrix}} & = & \nm{\begin{bmatrix} \nm{\sqrt{\cfrac{\displaystyle\sum_{j=1}^{\nEX} \Big(\muj{\hat{X}} - \mumu{\hat{X}}\Big)^2}{\nEX}}} & \nm{\sqrt{\cfrac{\displaystyle\sum_{j=1}^{\nEX} \Big(\sigmaj{\hat{X}} - \musigma{\hat{X}}\Big)^2}{\nEX}}} & \nm{\sqrt{\cfrac{\displaystyle\sum_{j=1}^{\nEX} \Big({\lvert \maxj{\hat{X}} \rvert} - \mumax{\hat{X}}\Big)^2}{\nEX}}} \end{bmatrix}^T} \label{eq:Sim_Metrics_std_mean_std_max} \\
\nm{\begin{bmatrix} \nm{\maxmu{\hat{X}}} \\ \nm{\maxsigma{\hat{X}}} \\ \nm{\maxmax{\hat{X}}} \end{bmatrix}} & = & \nm{\begin{bmatrix} \nm{\displaystyle\max_{j=1}^{\nEX} \ \lvert \muj{\hat{X}} \rvert} & \nm{\displaystyle\max_{j=1}^{\nEX} \ \lvert \sigmaj{\hat{X}} \rvert} & \nm{\displaystyle\max_{j=1}^{\nEX} \ \lvert \maxj{\hat{X}} \rvert} \end{bmatrix}^T} \label{eq:Sim_Metrics_max_mean_std_max} 
\end{eqnarray}

These metrics provide an overall view of the navigation system behavior under different circumstances, as they take into consideration different combinations of the stochastic parameters that define the two scenarios.


\subsubsection*{Aggregated Final State Metrics}

The aggregated metrics are adequate to evaluate the navigation system performance as long as each evaluation of \nm{f_{Xj} \lrp{t_n}} can be considered a realization of a stochastic process, which is the case for those variables that can be tracked (with more or less accuracy) without drift or accumulated errors by the navigation system. If a certain variable presents a drift, this is, the estimation error has a tendency to increase with time, then both the trajectory as well as the aggregated metrics are meaningless as the results depend on the execution length \nm{\tEND}. In this case, it makes sense to compute the \emph{aggregated final state metrics}, which rely on the final state value of every execution \nm{\seed}: 
\neweq{\begin{bmatrix} \nm{\muEND{\hat{X}}} \\ \nm{\sigmaEND{\hat{X}}} \\ \nm{\maxEND{\hat{X}}} \end{bmatrix} = \begin{bmatrix} \nm{\cfrac{\displaystyle\sum_{j=1}^{\nEX} \XENDj{\hat{X}}}{\nEX}} & \nm{\sqrt{\cfrac{\displaystyle\sum_{j=1}^{\nEX} \Big(\XENDj{\hat{X}} - \muEND{\hat{X}} \Big)^2}{\nEX}}} & \nm{\displaystyle\max_{j=1}^{\nEX} \ \lvert \XENDj{\hat{X}} \rvert} \end{bmatrix}^T}{eq:Sim_Metrics_mean_std_max_end} 


\subsubsection*{Time Aggregated Metrics}

The \emph{time aggregated metrics} consider that all evaluations of the error function \nm{f_{Xj}} for different runs at a given time instant \nm{t_n} together make up a random vector \nm{f_{Xn}} of size \nm{\nEX}, and then computes its mean \nm{\mun{\hat{X}}} and standard deviation \nm{\sigman{\hat{X}}}. They are primarily intended for graphical representation:
\begin{eqnarray}
\nm{\mun{\hat{X}}}    & = & \nm{\cfrac{\displaystyle\sum_{j=1}^{\nEX} f_{Xj} \lrp{t_n}}{\nEX}}\label{eq:Sim_Metrics_mean_s} \\
\nm{\sigman{\hat{X}}} & = & \nm{\sqrt{\cfrac{\displaystyle\sum_{j=1}^{\nEX} \Big(f_{Xj} \lrp{t_n} - \mun{\hat{X}}\Big)^2}{\nEX}}}\label{eq:Sim_Metrics_std_s}
\end{eqnarray}


\subsubsection*{Drift, Bounded, and Biased Estimations}

When evaluating the capability of a navigation system to estimate or track a given variable, it is necessary to determime whether the estimation experiences a \emph{drift}, this is, the estimation error grows with time, or not, in which case the estimation error never exceeds a certain limit and is hence \emph{bounded}. If drift is not present, the individual executions or runs can be ordered in accordance with their trajectory metrics, and the system estimation capabilities evaluated by means of the aggregated metrics. If drift can not be avoided, then the final state value constitutes the best indicator for each simulation run, while the system performance needs to be evaluated by means of the aggregated final state metrics.

The bounded estimation of a given variable is \emph{unbiased} or \emph{zero mean} if the aggregated metrics for the means (\nm{\mumu{\hat{X}}}, \nm{\sigmamu{\hat{X}}}, \nm{\maxmu{\hat{X}}}) are significantly smaller than both the standard deviations (\nm{\musigma{\hat{X}}}, \nm{\sigmasigma{\hat{X}}}, \nm{\maxsigma{\hat{X}}}) and the maximums (\nm{\mumax{\hat{X}}}, \nm{\sigmamax{\hat{X}}}, \nm{\maxmax{\hat{X}}}), and their ratios are more pronounced as the number of executions \nm{\nEX} grows. If this is not the case, the estimation is known as \emph{biased}. When drift is present, an unbiased or zero mean estimation is that in which the aggregated final state mean \nm{\muEND{\hat{X}}} is significantly smaller than both the standard deviation \nm{\sigmaEND{\hat{X}}} and the maximum \nm{\maxEND{\hat{X}}}. 
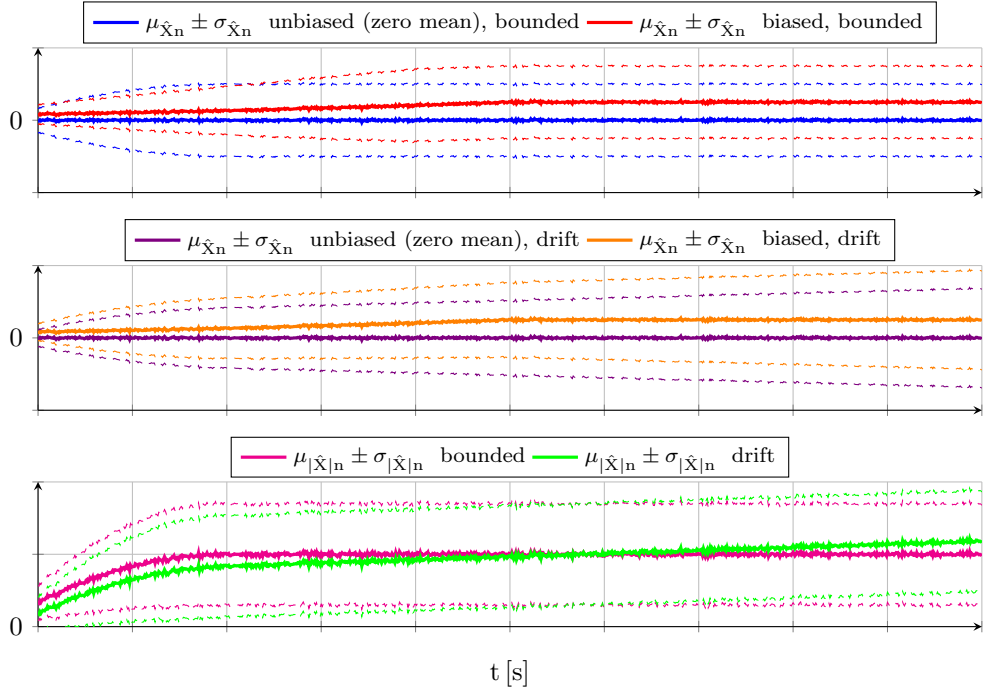
\begin{figure}[h]
\centering
\pgfplotsset{
	every axis legend/.append style={
		at={(0.50,1.03)},
		anchor=south,
	},
}
\begin{tikzpicture}
\begin{axis}[
cycle list={{blue,no markers,very thick},
			{red,no markers,very thick},
			{blue,dashed,no markers,thin},{blue,dashed,no markers,thin},
			{red,dashed,no markers,thin},{red,dashed,no markers,thin}},
width=14.0cm,
height=3.5cm,
xmin=0, xmax=1000, xtick={0,100,...,1000}, xticklabels={,,},
xmajorgrids,
ymin=-6, ymax=6, ytick={-6,0,6}, yticklabels={\empty,0,\empty},
ymajorgrids,
axis lines=left,
axis line style={-stealth},
legend entries={{\nm{\mun{\hat{X}} \pm \sigman{\hat{X}}} \ unbiased (zero mean), bounded},
                {\nm{\mun{\hat{X}} \pm \sigman{\hat{X}}} \ biased, bounded}},
legend columns=2,
legend style={font=\footnotesize},
legend cell align=left,
]
\draw [] (0.0,0.0) -- (1000.0,0.0); 
\pgfplotstableread{figs/metric_type.txt}\mytable
\addplot table [header=false, x index=0,y index=1]  {\mytable};
\addplot table [header=false, x index=0,y index=4]  {\mytable};
\addplot table [header=false, x index=0,y index=2]  {\mytable};
\addplot table [header=false, x index=0,y index=3]  {\mytable}; 
\addplot table [header=false, x index=0,y index=5]  {\mytable};
\addplot table [header=false, x index=0,y index=6]  {\mytable};
\end{axis}   
\end{tikzpicture}
\pgfplotsset{
	every axis legend/.append style={
		at={(0.50,1.03)},
		anchor=south,
	},
}
\begin{tikzpicture}
\begin{axis}[
cycle list={{violet,no markers,very thick},
			{orange,no markers,very thick},
			{violet,dashed,no markers,thin},{violet,dashed,no markers,thin},
			{orange,dashed,no markers,thin},{orange,dashed,no markers,thin}},
width=14.0cm,
height=3.5cm,
xmin=0, xmax=1000, xtick={0,100,...,1000}, xticklabels={,,},
xmajorgrids,
ymin=-6, ymax=6, ytick={-6,0,6}, yticklabels={\empty,0,\empty},
ymajorgrids,
axis lines=left,
axis line style={-stealth},
legend entries={{\nm{\mun{\hat{X}} \pm \sigman{\hat{X}}} \ unbiased (zero mean), drift},
                {\nm{\mun{\hat{X}} \pm \sigman{\hat{X}}} \ biased, drift}},
legend columns=2,
legend style={font=\footnotesize},
legend cell align=left,
]
\draw [] (0.0,0.0) -- (1000.0,0.0); 
\pgfplotstableread{figs/metric_type.txt}\mytable
\addplot table [header=false, x index=0,y index=7]  {\mytable};
\addplot table [header=false, x index=0,y index=10] {\mytable};
\addplot table [header=false, x index=0,y index=8]  {\mytable};
\addplot table [header=false, x index=0,y index=9]  {\mytable};
\addplot table [header=false, x index=0,y index=11] {\mytable};
\addplot table [header=false, x index=0,y index=12] {\mytable};
\end{axis}   
\end{tikzpicture}
\pgfplotsset{
	every axis legend/.append style={
		at={(0.50,1.03)},
		anchor=south,
	},
}
\begin{tikzpicture}
\begin{axis}[
cycle list={{magenta,no markers,very thick},
			{green,no markers,very thick},
			{magenta,dashed,no markers,thin},{magenta,dashed,no markers,thin},
			{green,dashed,no markers,thin},{green,dashed,no markers,thin}},
width=14.0cm,
height=3.5cm,
xmin=0, xmax=1000, xtick={0,100,...,1000}, xticklabels={,,},
xmajorgrids,
xlabel={\nm{t \lrsb{s}}},
ymin=0, ymax=6, ytick={0,3,6}, yticklabels={0,\empty,\empty},
ymajorgrids,
axis lines=left,
axis line style={-stealth},
legend entries={{\nm{\mun{\lvert \hat{X} \rvert} \pm \sigman{\lvert \hat{X} \rvert}} \ bounded},
                {\nm{\mun{\lvert \hat{X} \rvert} \pm \sigman{\lvert \hat{X} \rvert}} \ drift}},
legend columns=2,
legend style={font=\footnotesize},
legend cell align=left,
]
\pgfplotstableread{figs/metric_type.txt}\mytable
\addplot table [header=false, x index=0,y index=13]  {\mytable};
\addplot table [header=false, x index=0,y index=16]  {\mytable};
\addplot table [header=false, x index=0,y index=14]  {\mytable};
\addplot table [header=false, x index=0,y index=15]  {\mytable}; 
\addplot table [header=false, x index=0,y index=17]  {\mytable};
\addplot table [header=false, x index=0,y index=18]  {\mytable};
\end{axis}   
\end{tikzpicture}
\caption{Examples of time aggregated metrics results}
\label{fig:metric_type}
\end{figure}

If a variable corresponds to a scalar or vector component, this is, it can hold positive and negative values\footnote{Each of the body attitude Euler angles, and the geometric altitude, for example.}, its estimation can be either biased or unbiased, and can also drift or be bounded. A bias is generally indicative of a problem or lack of balance in either the estimation algorithm or the input data itself, while drift may caused by defects in the algorithm or lack of observability. Examples of the four possibilities are shown in the first two plots of figure \ref{fig:metric_type}.

On the other hand, if the variable corresponds with an absolute value \nm{< \lvert \cdot \rvert >} or the norm of a vector \nm{< \| \cdot \| >}, this is, it can only hold positive values\footnote{The attitude rotation vector and the horizontal position, for example.}, its estimation is always biased (unless it is perfect and the error is zero) and the main consideration is whether drift is present or the estimation is bounded. The third plot within figure \ref{fig:metric_type} shows these two possibilities.

The same classification can be applied to the individual estimations for a given run instead of the time aggregated metrics. It is common and often unavoidable that the individual estimations are biased or exhibit drift while these dissapear when aggregated. Although not ideal, this generally indicates the the source of the bias or drift is random and is external to the algorithm itself, originating within the observations or input data.

 
\section{Conclusions} \label{sec:Conclusions}

This article describes an stochastic high fidelity simulation of the flight of a fixed wing low SWaP autonomous UAV in turbulent and varying weather, for which the open-source \nm{\CC} implementation is available in \cite{Gallo2020_simulation}. The simulation is intended to test the behavior of different inertial, visual, and visual-inertial navigation systems in GNSS-Denied conditions, in which current systems experience an unbounded position drift or position error growth with time. The article describes the different simulation modules, defines two scenarios representative of GNSS-Denied conditions, and presents the metrics employed for the evaluation of the navigation performance. Future articles will make use of this simulation infrastructure to test novel GNSS-Denied navigation algorithms proposed by the author. 

\bibliographystyle{ieeetr}   
\bibliography{simulation}

\end{document}